\documentclass[conference]{IEEEtran}
\IEEEoverridecommandlockouts
\usepackage{xcolor}
\usepackage[dvipsnames]{xcolor}
\usepackage{amsmath}
\usepackage{graphicx}
\usepackage[table]{xcolor} 
\usepackage{booktabs}  
\usepackage{tabularx}  
\usepackage[utf8]{inputenc}
\usepackage{array}
\usepackage{tikz}
\usetikzlibrary{trees, positioning}
\usepackage{longtable}
\usepackage{caption}
\usepackage{subcaption}
\usepackage{multirow}
\usepackage{tabularx}
\usepackage{cuted}

\begin{document}

\bibliographystyle{IEEEtran}

\title{Color Models in Image Processing: A Review and Experimental Comparison
\thanks{Corresponding author: Muragul Muratbekova (e-mail: muragulm@gmail.com).}%
\thanks{This research was funded by the Science Committee of the Ministry of Science and Higher Education of the Republic of Kazakhstan (Grant No. AP22786412).}%
}

\author{
\IEEEauthorblockN{
Muragul Muratbekova\IEEEauthorrefmark{1},
Nuray Toganas\IEEEauthorrefmark{1},
Ayan Igali\IEEEauthorrefmark{1},
Maksat Shagyrov\IEEEauthorrefmark{1},
Elnara Kadyrgali\IEEEauthorrefmark{1},\\
Adilet Yerkin\IEEEauthorrefmark{1},
Pakizar Shamoi\IEEEauthorrefmark{1},  
}
\IEEEauthorblockA{\IEEEauthorrefmark{1}School of Information Technology and Engineering, Kazakh-British Technical University, Almaty, Kazakhstan}
}

\maketitle

\begin{abstract}
Color representation is essential in computer vision and human-computer interaction. There are multiple color models available. The choice of a suitable color model is critical for various applications.
This paper presents a review of color models and spaces, analyzing their theoretical foundations, computational properties, and practical applications. We explore traditional models such as RGB, CMYK, and YUV, perceptually uniform spaces like CIELAB and CIELUV, and fuzzy-based approaches as well. Additionally, we conduct a series of experiments to evaluate color models from various perspectives, like device dependency, chromatic consistency, and computational complexity. 
Our experimental results reveal gaps in existing color models and show that the HS* family is the most aligned with human perception. The review also identifies key strengths and limitations of different models and outlines open challenges and future directions
  This study provides a reference for researchers in image processing, perceptual computing, digital media, and any other color-related field.
\end{abstract}


%
%

\section*{Introduction}



Color perception is fundamental to image processing, computer vision, multimedia systems, visual communications, marketing, design, and human-computer interaction \cite{Gevers2008Color, Trémeau2008Color}. For example, in computer vision, representing and manipulating colors is crucial for various tasks, including object detection, classification, and scene understanding. It is a complex phenomenon, so various mathematical models have been developed to represent and manipulate color for different applications \cite{Burambekova2024Comparative}.

Despite the vast number of color models available, each has limitations in terms of perceptual accuracy, computational complexity, and device dependency.  Some of them are highly application-specific.  Understanding color science and applying the right one to design algorithms leads to better visual quality and reduced computational complexity \cite{Reinhard2009Color, Post1997Color}. The human visual system perceives colors differently from how they are represented in computers \cite{Taylor2020Representation}. Humans perceive color via a complex process involving neural \cite{Kim2020Neural}, contextual, and subjective factors like unique experience and prior knowledge \cite{Vandenbroucke2016Prior} that are not fully captured by computer color models. In the human brain, color perception involves a transition from processing the physical stimulus at the retina to a mental construct of color in higher visual areas \cite{Kim2020Neural}. This transition is not accounted for in digital color models using fixed numerical values.

Traditional models, such as RGB, are widely used but lack perceptual uniformity \cite{Shamoi2014Colorspace}. On the other hand, CIE and HS* family models improve perceptual consistency but are computationally expensive \cite{Paschos2001Perceptually, Mureika2005Fractal}. The need for optimized color models that balance accuracy and efficiency has become more critical nowadays with the rise of deep learning and computer vision applications. 


This paper aims to provide a comprehensive review and analysis of color models and spaces, including their mathematical foundations, computational properties, and real-world applications. We examine traditional models such as RGB and CMYK, perceptually uniform models like CIELAB and CIELUV, and recent advances in fuzzy color spaces, among others. Furthermore, we assess the efficiency of different models through empirical experiments to understand the strengths and weaknesses of each model in a real example in real settings. In addition, we highlight emerging trends in color representation. 

The contributions of this review are as follows:
\begin{itemize}
\item Comprehensive analysis of color models and spaces, including traditional (RGB, CMYK, YUV) and perceptually uniform (CIELAB, CIELUV) models, as well as fuzzy-based color models. 
\item Comparative evaluation of color models via empirical experiments. As a result, we identified trade-offs between processing speed, perceptual uniformity, and other aspects.

\item Identification of research gaps and future directions
\end{itemize}

The paper is organized as follows. Section I is this Introduction. Section II outlines the methodology for selecting relevant literature: sources, keywords, and inclusion criteria. Section III provides a literature review of various color models, including traditional, perceptually uniform, and fuzzy-based models, discussing their applications and limitations. Section IV presents experimental results comparing the efficiency of color models from various perspectives. Section V highlights gaps in the reviewed models and outlines open research challenges and directions for future work. Finally, Section VI provides the concluding remarks of this study.

\section*{Methods}

\begin{figure}
    \centering
    \includegraphics[width=\linewidth]{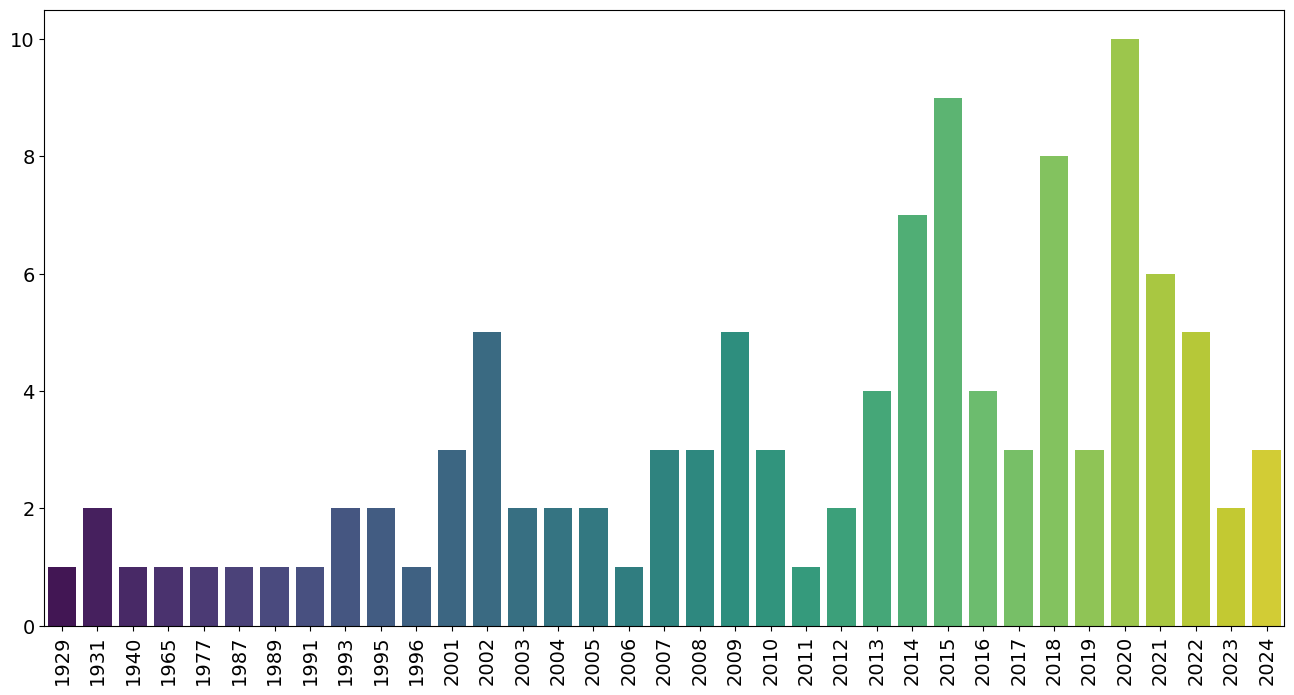}
    \caption{Number of Articles Published Per Year}
    \label{fig:number_articles_color}
\end{figure}

\begin{figure}
    \centering
    \begin{subfigure}[b]{0.5\textwidth}
        \centering
        \includegraphics[width=\linewidth]{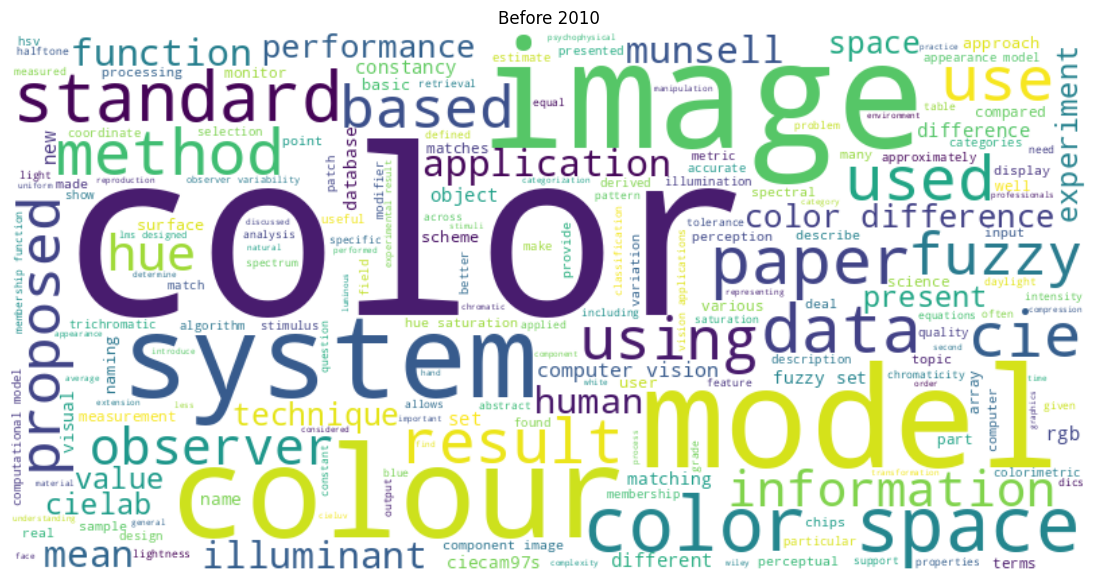}
        \caption{Before 2010}
    \end{subfigure}%
     \hfill
    \begin{subfigure}[b]{0.5\textwidth}
        \centering
        \includegraphics[width=\linewidth]{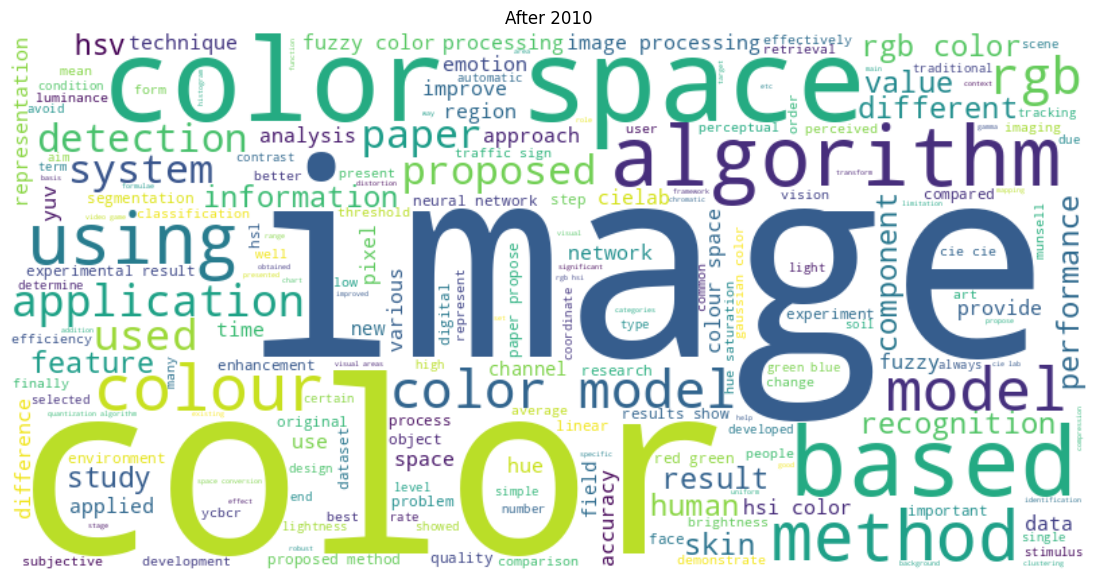}
        \caption{After 2010}
    \end{subfigure}
    \caption{Word Clouds of abstracts from publications in different time periods.}
    \label{abstract_wordcloud}
\end{figure}

\subsection*{Search strategy and selection criteria}
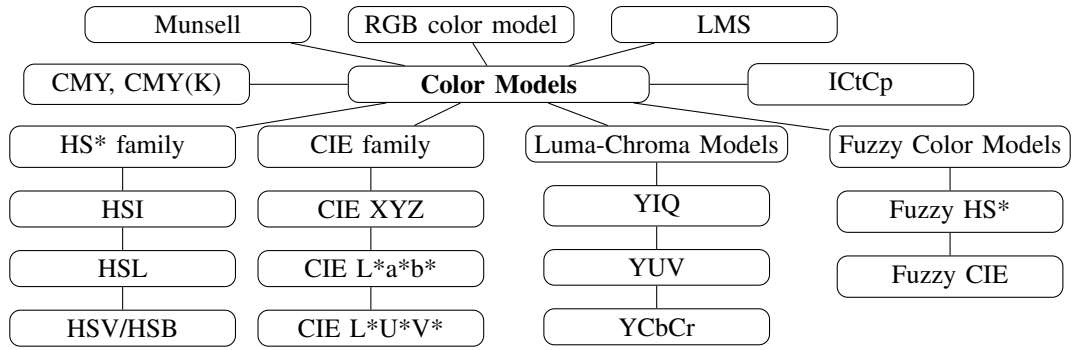
\begin{figure*}[ht!]
    \centering
    \begin{tikzpicture}[
        node distance=0.3cm and 1.3cm,
        every node/.style={draw, text centered, minimum width=3cm, rounded corners},
        main/.style={rectangle, draw, minimum width=4cm, font=\bfseries},
        branch/.style={rectangle, draw, minimum width=3cm}
    ]
    
    \node[main] (A) {Color Models};
    
    \node[branch, above=of A, xshift=-4cm] (B1) {Munsell};
    \node[branch, above=of A, xshift=-0.5cm] (B2) {RGB color model};
    \node[branch, above=of A, xshift=3cm] (D5) {LMS};
    \node[branch, right=of A] (D4) {ICtCp};
    \node[branch, left=of A] (B3) {CMY, CMY(K)};

    \node[branch, below=of A, xshift=-5cm] (B4) {HS* family};
    \node[branch, below=of A, xshift=-1.7cm] (B5) {CIE family};
    \node[branch, below=of A, xshift=2.1cm] (B6) {Luma-Chroma Models};
    \node[branch, below=of A, xshift=6cm] (B7) {Fuzzy Color Models};
    
    \node[branch, below=of B4] (C1) {HSI};
    \node[branch, below=of C1] (C2) {HSL};
    \node[branch, below=of C2] (C3) {HSV/HSB};
    
    \node[branch, below=of B5] (D1) {CIE XYZ};
    \node[branch, below=of D1] (D2) {CIE L*a*b*};
    \node[branch, below=of D2] (D3) {CIE L*U*V*};
    
    \node[branch, below=of B6] (E1) {YIQ};
    \node[branch, below=of E1] (E2) {YUV};
    \node[branch, below=of E2] (E3) {YCbCr};
    
    \node[branch, below=of B7] (F1) {Fuzzy HS*};
    \node[branch, below=of F1] (F2) {Fuzzy CIE};
        
    \draw (A) -- (B1);    
    \draw (A) -- (B2);    
    \draw (A) -- (B3);
    \draw (A) -- (D4);
    \draw (A) -- (D5);
    \draw (A) -- (B4);
    \draw (A) -- (B5);
    \draw (A) -- (B6);
    \draw (A) -- (B7);

    \draw (B4) -- (C1);
    \draw (C1) -- (C2);
    \draw (C2) -- (C3);

    \draw (B5) -- (D1);
    \draw (D1) -- (D2);
    \draw (D2) -- (D3);

    \draw (B6) -- (E1);
    \draw (E1) -- (E2);
    \draw (E2) -- (E3);

    \draw (B7) -- (F1);
    \draw (F1) -- (F2);

    \end{tikzpicture}
    \caption{Hierarchical Structure of Color Models}
    \label{fig:color_models}
\end{figure*}

The literature review covers publications from 1929 to 2025, with a primary focus on studies from 2001 to 2024 as shown in Fig. \ref{fig:number_articles_color}, ensuring the inclusion of both foundational works and recent advancements.  The sources include peer-reviewed journal articles, conference proceedings, books, and technical reports from Scopus- and Web of Science-indexed sources, including journals and conference proceedings from leading academic publishers such as IEEE, Springer, Elsevier, Wiley, ACM, and IET, among others. Articles were selected based on their relevance to such issues as fuzzy color spaces, image processing, color models, digital image enhancement, segmentation, and color-emotion associations. Priority was given to highly cited and influential works, including seminal papers. The review encompasses both theoretical and applied research, emphasizing contributions to fundamental theories, methodological advancements, and novel applications in color space analysis, fuzzy logic-based image processing, and color-emotion associations. By adopting this structured method, we prioritized the most commonly employed models, creating a robust basis for our study. Fig. \ref{fig:color_models} provides visual support.

\subsection*{Literature review analysis}
We analyzed abstracts of articles from literature review using NLP methods by illustrating a word cloud of keywords from the analyzed papers, as shown in Fig. \ref{abstract_wordcloud}. The comparison of abstracts before and after 2010 reveals a clear evolution in both tone and research priorities within color-model studies. Before 2010, there was a tendency towards theoretical and measurement-oriented works, dominated by terms such as 'standard', 'observer', 'illuminant', 'function', 'Munsell', and 'CIE'.
After 2010, the language shifted to an engineering and application-driven perspective. High frequency terms like 'algorithm', 'detection', 'recognition', 'network', 'segmentation' and 'skin' highlight a strong emphasis on automated image processing tasks and practical computer vision systems.
Additionally, post-2010 cloud adds 'skin', 'emotion',  'face', and 'human', demonstrating the growth in biometrics and affective computing. Moreover, after 2010, studies have increasingly integrated machine learning and neural network approaches for color-based image analysis, as demonstrated by terms such as 'algorithm', 'network', 'recognition', 'classification', 'segmentation', 'feature', and 'detection' in cloud computing.

\subsection*{Definitions}

The \textit{color model} is an abstract mathematical concept that describes how colors are represented numerically using sets of values (e.g. RGB, CMYK, HSV) \cite{Hasan2012, Yong2018A, Doncean2021ABSTRACT}.

The \textit{color space} is a specific implementation of a color model that defines the gamut of colors that can be displayed or printed, often tailored for devices or perceptual uniformity (e.g. sRGB, Adobe RGB, CIELAB, CAM16-UCS) \cite{Palus1998Representations, Briggs2020Colour, Doncean2021ABSTRACT}. The definition of a color space balances the choice of primary colors, signal noise, and the number of digital levels supported by a file type.

The term \textit{color system} is ambiguous and may refer to standardized color classification systems (e.g. Munsell, Pantone), color management workflows or, in some contexts, be used interchangeably with color models and color spaces.\cite{lammens1994computational}

\section*{Color Models}

\subsection*{Munsell}

\begin{figure*}[ht]
     \centering
     \begin{subfigure}[b]{0.45\linewidth}
         \centering
        \includegraphics[width=\linewidth]{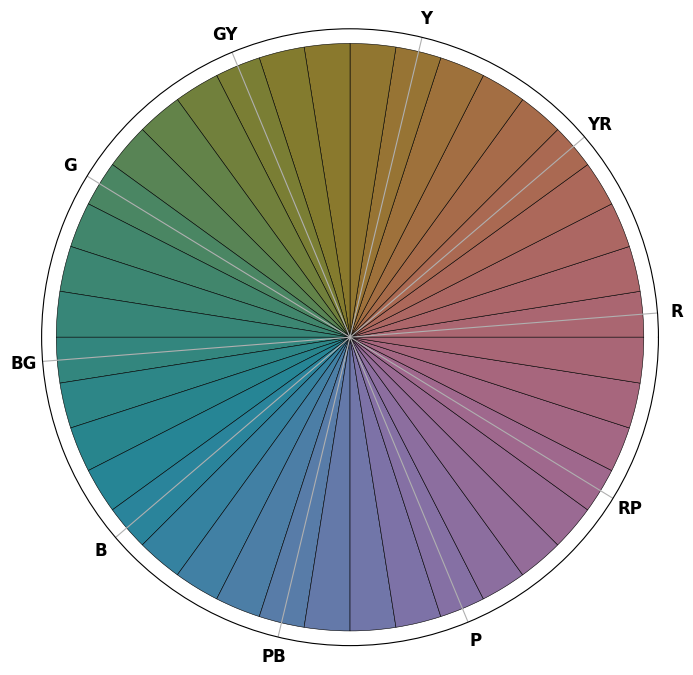}
        \caption{Munsell Hue Circle (Value=5, Chroma $\approx 6$)}
     \end{subfigure}
     \hfill
     \begin{subfigure}[b]{0.5\linewidth}
        \centering
        \includegraphics[width=\linewidth]{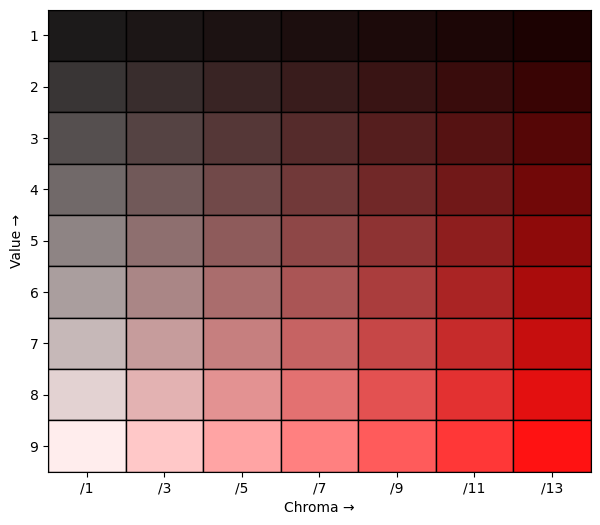}
        \caption{Munsell Value–Chroma Grid (Hue=Red)}
     \end{subfigure}
        \caption{Munsell color system.}
        \label{fig:munsell_components}
\end{figure*}

The Munsell Color System, developed by Albert H. Munsell in 1905, is one of the most widely accepted and practical color classification systems. It organizes colors within a three-dimensional model based on hue, chroma, and value, offering a precise and standardized method for describing and categorizing colors \cite{Munsell1905}. Fig. \ref{fig:munsell_components} illustrates the hue circle at a fixed value (5) and chroma (6), showing the ten primary hue families and their subdivisions.

\subsubsection*{1. Hue (Color Tone)}

Hue represents the dominant wavelength of reflected light, which determines a color's tonality. In Munsell’s spherical color space, hues are arranged along the equator in a circular formation, creating the "color wheel." Primary colors (red, yellow, and blue) and secondary colors (orange, green, and violet) are systematically positioned according to their spectral relationships.

\subsubsection*{2. Chroma (Saturation)}

Chroma measures the intensity or purity of a color, reflecting the degree of selective light reflection. In the Munsell model, chroma increases as a color moves outward from the center of the sphere, becoming more vivid and saturated. Colors with low chroma appear weak or muted, whereas those with high chroma are vivid and rich.

\subsubsection*{3. Value (Brightness)}

Value represents the lightness or darkness of a color, determined by the proportion of white, gray, or black it contains. In Munsell’s color sphere, white is positioned at the top, black at the bottom, and gray along the vertical axis, illustrating the gradient of brightness. A higher value corresponds to lighter colors, whereas a lower value indicates darker shades.

\subsubsection*{4. Attribute Dependency}
The Munsell system is designed to ensure perceptual independence between hue, chroma, and value, meaning changes in one attribute should not inherently affect the others. However, some interactions exist. High-chroma colors may appear slightly lighter than low-chroma ones with the same Munsell value due to increased light reflection, creating a sense of brightness. Additionally, certain hues achieve higher chroma levels than others due to limitations in pigment and display. For example, yellow can reach a higher chroma than blue before shifting in hue or value. Similarly, yellows tend to have higher values than purples, as they reflect more light, making them appear brighter even at the same chroma level.

One of the Munsell system’s strengths is its clear separation of chroma and value, ensuring precise color categorization. Chroma is measured radially from the sphere’s center, controlling intensity without affecting brightness, while value is measured along the vertical axis, allowing systematic lightness comparisons regardless of hue or chroma. This structure enables accurate color classification, matching, and reproduction across various applications.

The system was refined through extensive visual experiments, including work by the Optical Society of America in the late 1930s \cite{Kheng2005ColorSA}. It is widely applied in fields such as archaeology, geology, and anthropology for identifying colored surfaces and soils \cite{Stanco11}. However, its subjective nature can lead to errors. To address this, researchers have developed automatic methods for color detection in digital images of archaeological pottery using the Munsell system \cite{Stanco11}.

Research on the Munsell system and human color perception provides valuable insights into color constancy and unique hue selection. Color constancy—the ability to recognize surface colors under varying lighting—is more reliable for real surfaces than digitally rendered ones, with the highest consistency observed around prototypical hues \cite{Olkkonen2009}. Studies using Munsell chips indicate intra-observer variability accounts for approximately 15\% of inter-observer variability, with no significant gender differences when all chips are presented simultaneously \cite{Hinks:07}. In controlled environments, surface-color matches exhibit stronger color constancy than hue-saturation matches \cite{Arend:91}. Additionally, human color-matching functions and cone sensitivities provide nearly identical information for estimating reflectance spectra, with results closely aligning with Munsell reflectance measurements under D65 illumination \cite{Romney2002}. These findings enhance our understanding of human color perception and its connection to physical color attributes.

\subsection*{RGB color model}

Color perception is facilitated by the transmission of three signals to the brain, reflecting nature’s tendency toward minimalist processing. The RGB color model corresponds to the biological processing of colors in the human visual system \cite{Pascale2003, Loesdau2014}.

\begin{figure}
     \centering
     \begin{subfigure}[b]{0.49\linewidth}
         \centering
        \includegraphics[width=\linewidth]{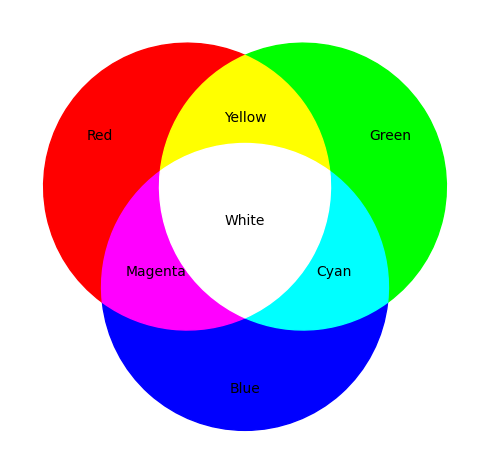}
        \caption{RGB (additive)}
        \label{fig:rgb}
     \end{subfigure}
     \begin{subfigure}[b]{0.49\linewidth}
        \centering
        \includegraphics[width=\linewidth]{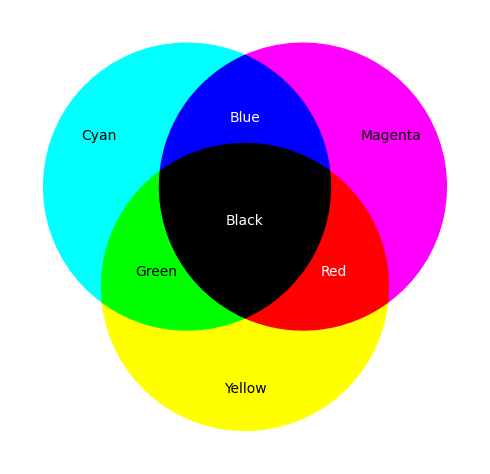}
        \caption{CMY (subtractive)}
        \label{fig:cmy}
     \end{subfigure}
        \caption{Schematic view of the RGB and CMY color models.}
\end{figure}

RGB is an additive color model in which red, green, and blue light are combined in varying proportions to generate a broad spectrum of colors. The schematic view can be observed in Fig. \ref{fig:rgb}. It is the most commonly used color space, particularly in image processing systems, computer graphics, and multimedia applications. 
Each channel ranges from 0 to 255, and the gamut can be visually represented as a cube. However, the strong correlation between R, G, and B is the main reason why RGB is not the best choice for many applications.\cite{Sharma2015, Ketenci2013}

Since there is no single RGB color space suitable for all imaging needs, it is vital to first understand the overall digital image color workflow before examining the characteristics and usage of different RGB color spaces. There are various RGB color spaces (see Table \ref{rgbtable}), such as sRGB, Adobe RGB, Apple RGB, CIE RGB, ColorMatch RGB, HDTV RGB, NTSC RGB, PAL/SECAM RGB, SGI RGB, SMPTE-C, and SMPTE-240M RGB, as well as standard RGB color spaces like ISO RGB, ROMM RGB, and video RGB spaces (EBU, ITU-R BT.709)\cite{Pascale2003, Ssstrunk1999StandardRC}.

\begin{table*}[t]
\caption{Overview of RGB color spaces}
\label{rgbtable}
\centering
\renewcommand{\arraystretch}{1.2}
\begin{tabularx}{\textwidth}{|p{0.15\textwidth}|X|}
\hline
\textbf{Name} & \textbf{Description} \\
\hline
sRGB & sRGB is defined in IEC 61966-2-1 as the default color space for multimedia. 
It is based on the characteristics of a reference CRT display, establishing a link between 8-bit sRGB values and CIE 1931 XYZ \cite{Ssstrunk1999StandardRC}. 
HP and Microsoft proposed integrating sRGB into operating systems, HP products, and the Internet. This device-independent color space ensures high quality, backward compatibility, and minimal system load \cite{Anderson1996ProposalFA}. 
sRGB ensures consistent color display on screens, the web, and in print. Monitors reproduce about 97\% of their colors, so prints closely match the on-screen image. Browsers don’t alter colors, and with a calibrated monitor, the difference between screen and print is minimal \cite{srgb}. \\
\hline
Adobe RGB & Photoshop Working Space. In Photoshop 5, Adobe introduced a device-independent working space, enabling image editing that is not tied to a specific monitor.
Adobe RGB 98, based on the SMPTE-240M standard, was designed for pre-press work with a wide color gamut. However, it includes colors that standard CMYK printers cannot reproduce, so considering the target output gamut is essential \cite{Ssstrunk1999StandardRC}. 
Although many colors, particularly in the green range, cannot be reproduced using the SWOP standard, modern technologies such as Pantone Hexachrome make effective use of this gamut \cite{Pascale2003}. 
Regular monitors display 75\% of Adobe RGB, while professional ones show up to 98\%. \\
\hline
ISO RGB & ISO RGB is the only unrendered RGB color space currently in the standardization process (ISO 17321). It is designed for evaluating digital camera color performance, with sensor-to-ISO RGB transformations defined in the standard.
Unlike CIE XYZ, ISO RGB is optimized for digital cameras, ensuring accurate white point conversion and allowing transformation into CIE XYZ through matrix conversion \cite{Ssstrunk1999StandardRC}. \\
\hline
Apple RGB & Apple RGB is based on the 13" Apple RGB monitor and was the standard in publishing software like Photoshop and Illustrator. Its gamut is similar to sRGB, but it remains prevalent in legacy desktop publishing files \cite{Ssstrunk1999StandardRC}.
Apple RGB uses a non-unity display LUT gamma, which is compensated by the file encoding gamma \cite{Pascale2003}. \\
\hline
ROMM RGB & ROMM RGB, developed by Kodak, is a color space for image editing after rendering. 
With a D50 white point, standard for printing, ROMM RGB has the widest gamut among rendered RGB spaces. However, this results in unused values in printing and potential artifacts with 8-bit quantization. To ensure precise color reproduction, 12- and 16-bit formats are supported. ROMM RGB is available as a working space in Adobe Photoshop 5 \cite{Ssstrunk1999StandardRC}. \\
\hline
\end{tabularx}
\end{table*}

\subsubsection*{Linear RGB and Non-linear R'G'B' color spaces}
When an image acquisition system, such as a video camera, captures an object's image, it receives the linear light emitted by the object. 
The camera then applies gamma correction to convert the linear RGB intensities into non-linear RGB signals.{\cite{Poynton1996ATI, Plataniotis2000}}
Digital image pixels acquired from an object are stored as non-linear R'G'B' values, typically in the 0–255 range. A color image pixel requires three bytes—one for each component (R', G', B'). These values are stored in image files on computers and used in image processing tasks. The transformation to non-linear R'G'B' values in the range (0, 1) from linear RGB values in the range (0, 1) is defined by:
\begin{equation}
R' =
\begin{cases}
4.5R, & \text{if } R \leq 0.018 \\
\frac{1.099R^{\frac{1}{\gamma_C}} - 0.099}{1.099}, & \text{otherwise}
\end{cases}
\end{equation}

\begin{equation}
G' =
\begin{cases}
4.5G, & \text{if } G \leq 0.018 \\
\frac{1.099G^{\frac{1}{\gamma_C}} - 0.099}{1.099}, & \text{otherwise}
\end{cases}
\end{equation}

\begin{equation}
B' =
\begin{cases}
4.5B, & \text{if } B \leq 0.018 \\
\frac{1.099B^{\frac{1}{\gamma_C}} - 0.099}{1.099}, & \text{otherwise}
\end{cases}
\end{equation}
where $\gamma_C$ is known as the gamma factor of the camera or the acquisition device. \cite{Poynton1996ATI, Plataniotis2000}

\begin{figure*}[ht]
    \centering
    \includegraphics[width=\linewidth]{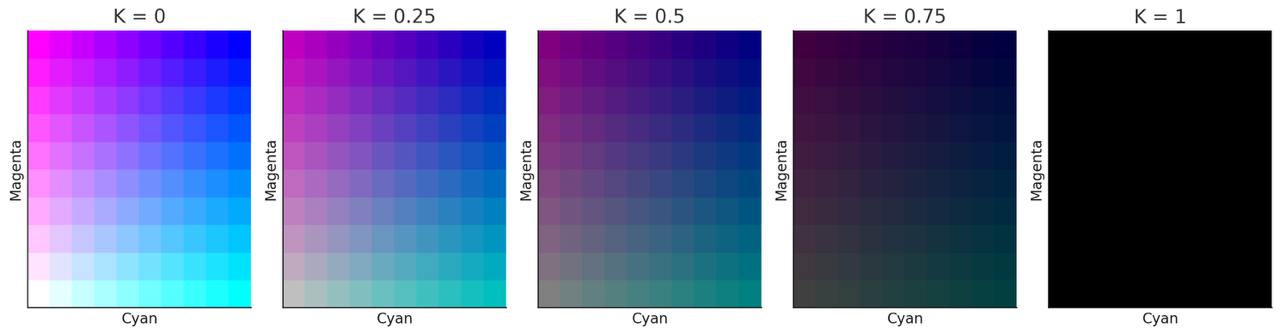}
    \caption{2D cross-sections of the CMYK color space in the Cyan–Magenta plane at varying levels of the K (Black) component.}
    \label{fig:cmyk_slice}
\end{figure*}

\subsection*{CMY(K) color model} 
CMY is a subtractive, device-dependent color model where cyan, magenta, and yellow inks are applied to a white surface, absorbing light to create the desired color. Components and their relation can be seen in Fig. \ref{fig:cmy}. In the CMY model, black is produced by mixing C, M, and Y, but it does not generate a pure black. To address this, an additional K (Black) component was introduced, enabling deep black printing.
Color components: C, M, Y range from (0-255), K is either (0 or 1)\cite{Kour2015} 

In printing, white paper is typically used, which prevents light from passing through. Therefore, in industrial color printing, CMYK colors (cyan, magenta, yellow, and black) are applied sequentially. Shades are formed similarly to mixing paints: for example, combining blue and yellow creates green, with its intensity depending on the ratio of water and ink.


\subsubsection*{CMYK color space}
The CMYK color space is a subtractive model used in printing.\cite{Sawicki2015}
\begin{itemize}
    \item {Cyan} – a light, cool shade of blue.
    \item {Magenta} – a mix of red and blue, similar to crimson and amaranth.
    \item {Yellow} – slightly paler than standard yellow, created by mixing red and green.
    \item {Black} – used to save colored ink and ensure printing stability.
\end{itemize}
To represent the 4D CMYK color model in a comprehensible way, Fig. \ref{fig:cmyk_slice} presents 2D cross-sections along the Cyan–Magenta plane, with different fixed values of the K component. This illustrates how the color space evolves as the black component increases. Each slice corresponds to a fixed K value, while Yellow is held constant.


The following equations give the transformation from RGB to CMYK \cite{gonzalez2018digital}:
\begin{align}
C &= \frac{1 - \frac{R}{255} - K}{1 - K} \quad \text{if } (1 - K) \neq 0 \text{, otherwise } 0 \\
M &= \frac{1 - \frac{G}{255} - K}{1 - K} \quad \text{if } (1 - K) \neq 0 \text{, otherwise } 0 \\
Y &= \frac{1 - \frac{B}{255} - K}{1 - K} \quad \text{if } (1 - K) \neq 0 \text{, otherwise } 0 \\
K &= 1 - \max\left(\frac{R}{255}, \frac{G}{255}, \frac{B}{255}\right)
\end{align}

The reverse transformation from the CMYK color model to the RGB color model:
\[
R = 255 \cdot (1 - C) \cdot (1 - K)
\]
\[
G = 255 \cdot (1 - M) \cdot (1 - K)
\]
\[
B = 255 \cdot (1 - Y) \cdot (1 - K)
\]
Where:
- \(C, M, Y, K\) are the components of Cyan, Magenta, Yellow, and Black (ranging from 0 to 1);
- \(R, G, B\) are the components of Red, Green, and Blue (ranging from 0 to 255).
If \(C, M, Y\) values are provided in the range [0, 255], they must first be normalized:
\[
C_{norm} = \frac{C}{255}, \quad M_{norm} = \frac{M}{255}, \quad Y_{norm} = \frac{Y}{255}
\]
Then, use the normalized values in the main formulas.



\subsection*{HS* family}

\begin{figure*}[ht]
     \centering
     \begin{subfigure}[b]{0.3\linewidth}
         \centering
        \includegraphics[width=\linewidth]{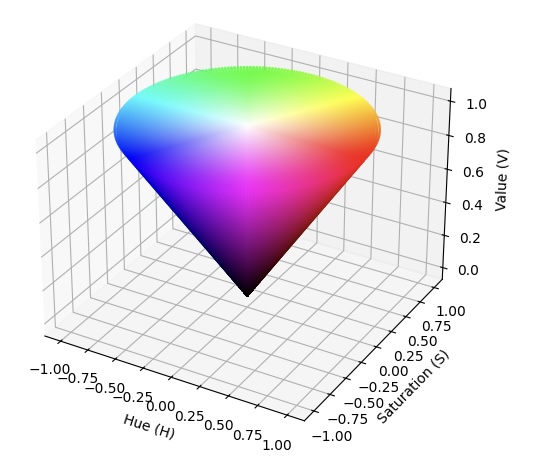}
        \caption{HSV Cone}
        \label{fig:hsv}
     \end{subfigure}
     \hfill
     \begin{subfigure}[b]{0.3\linewidth}
        \centering
        \includegraphics[width=\linewidth]{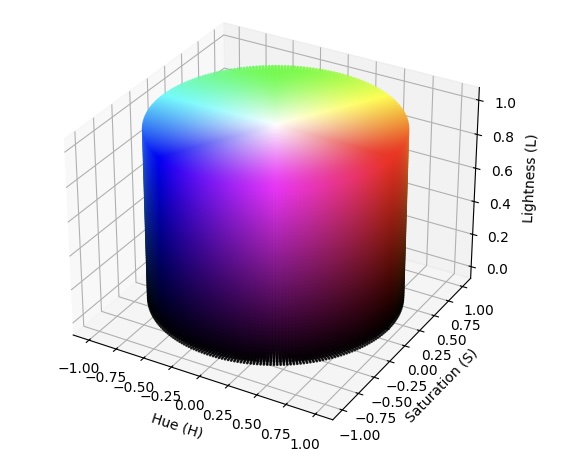}
        \caption{HSL Cylinder}
        \label{fig:hsl}
     \end{subfigure}
     \hfill
     \begin{subfigure}[b]{0.3\linewidth}
        \centering
        \includegraphics[width=\linewidth]{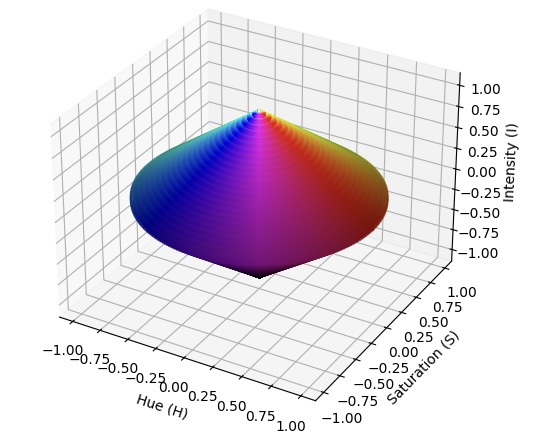}
        \caption{HSI Double Cone}
        \label{fig:hsi}
     \end{subfigure}
        \caption{HS-based color gamut visualized in (a) HSV cone, (b) HSL cylinder, and (c) HSI double cone geometries. The distributions illustrate how hue, saturation, and brightness/lightness/intensity are structured in each model.}
\end{figure*}

The HS* family has emerged as a critical tool in various fields requiring subjective and perceptual color interpretation, such as image processing, computer vision, and human-computer interaction. Unlike the RGB (Red, Green, Blue) color model, which is based on the additive mixing of primary colors, the HS* models are designed to represent colors in a way that is more aligned with human visual perception. They decompose colors into different components (hue, saturation, and a brightness-related component).
\begin{itemize}
    \item \textbf{Hue (H)}: The hue represents the type of color, such as red, green, or blue, and is defined by the angle in a cylindrical coordinate system. Hue is directly correlated with the wavelength of light, ranging from 0\textdegree{} to 360\textdegree{}, where specific colors are positioned along this circular dimension (e.g., red at 0\textdegree{}, green at 120\textdegree{}, and blue at 240\textdegree{}). This direct correspondence to the wavelength makes the model inherently more aligned with human color perception \cite{Barranco2005, Wang2014}. While the formula for H remains essentially the same in all the models, the way the other components (Saturation, Lightness, Brightness, or Intensity) affect the value of H can vary.
    \item \textbf{Saturation (S)}: Saturation indicates the purity of the color, ranging from 0 (completely desaturated or gray) to 1 (fully saturated, vibrant color). High saturation values correspond to pure, vivid colors, while lower values represent colors that are washed out or dull \cite{Barranco2005, Wang2014}. This property makes the HS* models particularly useful for tasks where the intensity of color needs to be adjusted without altering its hue.
\end{itemize}
The Hue and Saturation components are generally defined similarly across these color models, but their interpretations and calculations can differ slightly due to how the third component (lightness, intensity, brightness, or value) is handled.

Human vision processes colors in terms of hue, saturation, and a brightness-related component, making these models more intuitive than the Cartesian RGB model \cite{Ma2018, Zhi2020}. This perceptual alignment simplifies tasks requiring subjective color interpretation, such as image enhancement and emotional analysis. Additionally, transforming images into HS* color spaces enables the independent manipulation of components (e.g., brightness adjustments without altering hue or saturation), thereby preserving the overall color structure and reducing the risk of distortion \cite{Cui2022}.

Despite these advantages, the HS* models have limitations. The geometric representation, such as a cone or cylinder, introduces distortions in color space, particularly at low saturation and brightness levels. This can lead to inaccuracies in applications that require high precision \cite{Cao2019}. Computational overhead due to non-linear transformations, especially during RGB-to-HS* conversions, can also challenge real-time applications \cite{Zhi2020, Yu2023}.

\subsubsection*{HSI}
The HSI color model describes colors through three components: Hue (H), Saturation (S), and Intensity (I). \textbf{Intensity} measures the color's lightness, ranging from 0 (black) to 1 (white), though in digital applications, it is often scaled to [0,255].  Intensity is independent of hue and saturation, allowing brightness adjustment without changing the overall color tone \cite{Garg2022}.
The HSI model’s geometric representation is often described as a cone (Fig. \ref{fig:hsi}), with the intensity along the central axis, saturation increasing radially from the center, and hue spanning around the angular dimension. This structure provides a direct, perceptually relevant framework for color processing \cite{Garg2022}.

The following equations give the transformation from RGB to HSI:
\begin{equation}
H = \cos^{-1} \left( \frac{(R - G) + (R - B)}{2 \sqrt{(R - G)^2 + (R - B)(G - B)}} \right)
\end{equation}

\begin{equation}
S = 1 - \frac{\min(R, G, B)}{I}
\end{equation}

\begin{equation}
I = \frac{R + G + B}{3}
\end{equation}

The reverse transformation from HSI to RGB is computed using the following formulas:
\begin{equation}
R = I \left(1 + \frac{S \cos H}{\cos (60^\circ - H)} \right)
\end{equation}

\begin{equation}
G = I \left(1 + \frac{S (1 - \cos H / \cos (60^\circ - H))}{\cos (60^\circ - H)} \right)
\end{equation}

\begin{equation}
B = 3I - (R + G)
\end{equation}

These formulas are derived from standard color space conversion methods and are presented in \cite{winkler2005digital}.

The HSI model’s design reproduces how humans perceive and describe color. Human vision processes colors in terms of hue, saturation, and intensity, making HSI a more intuitive model for tasks requiring subjective color interpretation. This perceptual alignment has made it the model of choice in various applications, such as image enhancement, video game design, and emotional analysis, where human perception of color is crucial \cite{Ma2018, Zhi2020}.

While the HSI model has numerous advantages, it also has limitations. Under low saturation or intensity conditions, the hue component becomes unreliable, leading to inaccuracies in color representation \cite{Barranco2005, Geslin2016}. This limitation can affect applications requiring high precision, such as medical imaging and scientific analysis. Additionally, the conversion between RGB and HSI color spaces is computationally complex, involving non-linear transformations, which can be a challenge for resource-constrained environments \cite{Zhi2020, Yu2023}.

\subsubsection*{HSL}

The Hue, Saturation, and Lightness (HSL) color model is a cylindrical-coordinate representation designed to approximate human visual perception more intuitively than the Cartesian RGB model \cite{Kalist2015}. Visual representation of this model can be observed in Fig. \ref{fig:hsl}. \textbf{Lightness} measures the relative brightness of a color, ranging from 0\% (black) to 100\% (white). This component is independent of hue and saturation, making it particularly useful for applications like low-light image enhancement. Changing the Lightness channel does not alter the image's color information, allowing for enhanced clarity without over-enhancement or unnatural distortions. 

The transformation equations from RGB to HSL are given below.

\begin{strip}
\begin{equation}
H =
\begin{cases}
0^\circ, & \text{if } \max(R, G, B) = \min(R, G, B) \\
60^\circ \times \left( \dfrac{G - B}{\max(R, G, B) - \min(R, G, B)} \bmod 6 \right), & \text{if } \max(R, G, B) = R \\
60^\circ \times \left( \dfrac{B - R}{\max(R, G, B) - \min(R, G, B)} + 2 \right), & \text{if } \max(R, G, B) = G \\
60^\circ \times \left( \dfrac{R - G}{\max(R, G, B) - \min(R, G, B)} + 4 \right), & \text{if } \max(R, G, B) = B
\end{cases}
\end{equation}
\end{strip}

\begin{equation}
S =
\begin{cases}
0, & \text{if } \max(R, G, B) = \min(R, G, B) \\
\frac{\max(R, G, B) - \min(R, G, B)}{1 - |2L - 1|}, & \text{otherwise}
\end{cases}
\end{equation}

\begin{equation}
L = \frac{\max(R, G, B) + \min(R, G, B)}{2}
\end{equation}

The reverse transformation follows. The first step is to calculate the C, X, and m parameters using the following formulas.
\begin{equation}
C = (1 - |2L - 1|) \times S
\end{equation}

\begin{equation}
X = C \times (1 - |(H / 60^\circ) \mod 2 - 1|)
\end{equation}

\begin{equation}
m = L - C / 2
\end{equation}

RGB values are then computed depending on H:
\begin{equation}
(R, G, B) =
\begin{cases}
(C, X, 0), & 0^\circ \leq H < 60^\circ \\
(X, C, 0), & 60^\circ \leq H < 120^\circ \\
(0, C, X), & 120^\circ \leq H < 180^\circ \\
(0, X, C), & 180^\circ \leq H < 240^\circ \\
(X, 0, C), & 240^\circ \leq H < 300^\circ \\
(C, 0, X), & 300^\circ \leq H < 360^\circ
\end{cases}
\end{equation}

\begin{equation}
R = (R + m), \quad G = (G + m), \quad B = (B + m)
\end{equation}

These formulas are presented in \cite{poynton2012digital}.

HSL is designed to be more human-consistent than models like YCbCr and has a wider illuminance range than HSV, making it effective for applications requiring controlled color adjustments, segmentation, and image enhancement \cite{Garg2022, Kalist2015}.

Similar to other HS* models, the HSL model provides a more human-friendly description of color. Still, it also suffers from poor perceptual uniformity and uneven distribution of colors in its space, making it unsuitable for precise computational or perceptual tasks.

\subsubsection*{HSV/HSB}
The HSV (Hue, Saturation, Value) color model, often interchangeably referred to as HSB (Hue, Saturation, Brightness), provides an intuitive representation of color based on human perception. \textbf{Value (V)} indicates the brightness of the color, ranging from 0 (black) to 1 (full brightness) \cite{Cao2019, Zhang2018}.

Geometrically, the HSV color space is represented as an inverted cone or cylinder (Fig. \ref{fig:hsv}). The hue is depicted as the angle around the vertical axis, saturation as the radial distance from the center, and value as the vertical height. This design facilitates a visual understanding of color variations and their relationships \cite{Cao2019}.

The transformation equations from RGB to HSV are provided below.
\begin{strip}
\begin{equation}
H =
\begin{cases}
0^\circ, & \text{if } \max(R, G, B) = \min(R, G, B) \\
60^\circ \times \left( \dfrac{G - B}{V - \min(R, G, B)} \bmod 6 \right), & \text{if } V = R \\
60^\circ \times \left( \dfrac{B - R}{V - \min(R, G, B)} + 2 \right), & \text{if } V = G \\
60^\circ \times \left( \dfrac{R - G}{V - \min(R, G, B)} + 4 \right), & \text{if } V = B
\end{cases}
\end{equation}
\end{strip}

\begin{equation}
S =
\begin{cases}
0, & \text{if } V = 0 \\
\frac{V - \min(R, G, B)}{V}, & \text{otherwise}
\end{cases}
\end{equation}

\begin{equation}
V = \max(R, G, B)
\end{equation}

The inverse transformation is as follows.
\begin{equation}
C = V \times S
\end{equation}

\begin{equation}
X = C \times (1 - |(H / 60^\circ) \mod 2 - 1|)
\end{equation}

\begin{equation}
m = V - C
\end{equation}

The RGB components are determined based on H:
\begin{equation}
(R, G, B) = 
\begin{cases}
(C, X, 0), & 0^\circ \leq H < 60^\circ \\
(X, C, 0), & 60^\circ \leq H < 120^\circ \\
(0, C, X), & 120^\circ \leq H < 180^\circ \\
(0, X, C), & 180^\circ \leq H < 240^\circ \\
(X, 0, C), & 240^\circ \leq H < 300^\circ \\
(C, 0, X), & 300^\circ \leq H < 360^\circ
\end{cases}
\end{equation}

\begin{equation}
R = (R + m), \quad G = (G + m), \quad B = (B + m)
\end{equation}

These conversion formulas are taken from \cite{gonzalez2018digital}.

HSV is widely used in applications like color-based segmentation, visibility enhancement, and traffic sign detection due to its ability to separate chromatic content from brightness, preserve naturalness, and offer computational simplicity under varying lighting conditions \cite{Cao2019, Zhang2018, Yang2015}. Compared to RGB, HSV offers improved consistency under varying illumination. Similarly, it is computationally simpler than the HSI model while achieving comparable results in most practical scenarios \cite{Cao2019}.

Although widely used for intuitive representation of colors, the HSV/HSB model lacks perceptual uniformity, where equal numerical changes do not correspond to equal perceptual differences. It is also highly sensitive to variations in illumination, which limits its reliability in rigorous color analysis.

\subsection*{CIE family} 
CIE family color spaces are a set of standardized color spaces developed by the International Commission on Illumination \footnote{https://cie.co.at/} (CIE) to provide a consistent framework for color measurement and communication across different devices and applications \cite{Kang}.  CIE models describe how the visible spectrum relates to human color perception, incorporating a "standard observer"—a static representation of typical human color vision \cite{Smith1931}. These spaces are device-independent, i.e., not tied to any particular device's characteristics. Color spaces of the CIE family are popular for accurate color reproduction and comparison.

CIE family color spaces represent colors in a three-dimensional space using three key parameters \cite{Hasan2012}:
\begin{itemize} \item L* – Represents the lightness of a color, ranging from 0 (pure black) to 100 (pure white).
\item A* – Defines chromatic variation along the green-red axis, where negative values correspond to green and positive values to red.
\item B* – Represents the hue component on the blue-yellow axis, with negative values indicating blue and positive values indicating yellow.
\end{itemize}

The CIE 1931 color space was developed from Wright-Guild experimental data \cite{WWright_1929, 1931}, establishing foundational colorimetry principles still relevant today \cite{Fairman1997How}. The experimental results led to the CIE RGB color space, from which the CIE XYZ color space was derived. Over the years, the CIE has introduced various models, improving color difference and matching \cite{Brainard2020Colorimetry}.

CIE models are important in colorimetry, digital cameras, and color reproduction; e.g., smartphones rely on the CIE color system for accurate reproduction, involving optics, color filters, and sensors \cite{Kriss2015Color}. CIE models are also often used in image processing tasks, e.g., image enhancement. Enhancing an image involves methods like adjusting illumination standards, saturation, and white balance \cite{Florin2008A}. 

\begin{figure*}
     \centering
     \begin{subfigure}[b]{0.3\textwidth}
         \centering
        \includegraphics[width=\linewidth]{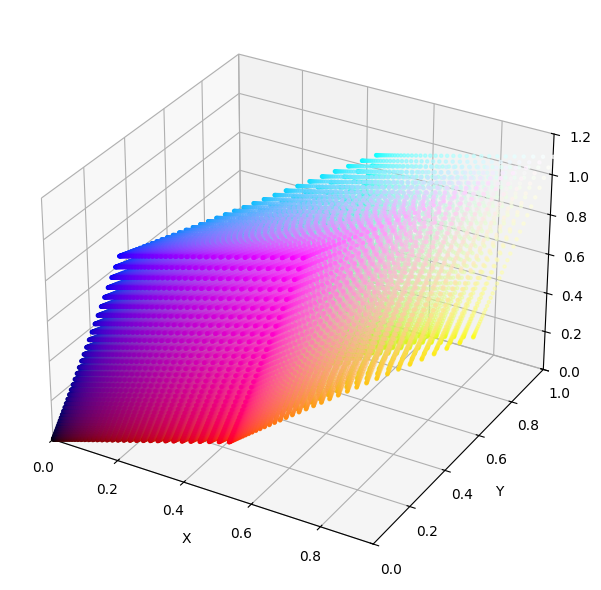}
        \caption{CIE XYZ}
        \label{fig:xyz}
     \end{subfigure}
     \hfill
     \begin{subfigure}[b]{0.3\textwidth}
        \centering
        \includegraphics[width=\linewidth]{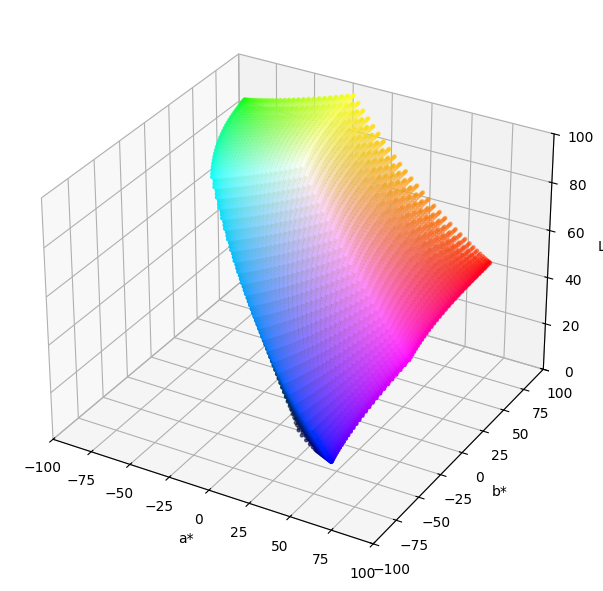}
        \caption{CIE L*a*b*}
        \label{fig:lab}
     \end{subfigure}
     \hfill
     \begin{subfigure}[b]{0.3\textwidth}
        \centering
        \includegraphics[width=\linewidth]{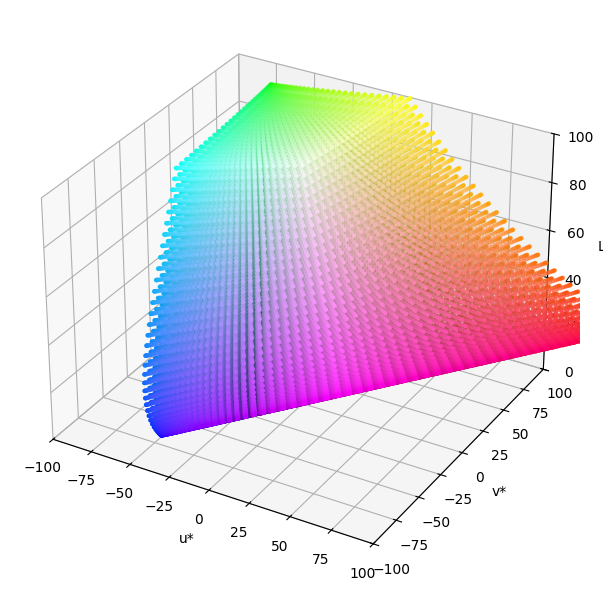}
        \caption{CIE L*u*v*}
        \label{fig:luv}
     \end{subfigure}
        \caption{RGB gamut visualized in (a) CIE XYZ, (b) CIE L*a*b*, and (c) CIE L*u*v* color spaces. The point distributions reveal how perceptual differences between colors vary across models.}
        \label{fig:cie_components}
\end{figure*}

As shown in Fig. \ref{fig:cie_components}, the RGB gamut appears differently across the CIE XYZ, CIE L*a*b*, and CIE L*u*v* color spaces.

\subsubsection*{CIE XYZ}
The CIEXYZ color space is the foundation of CIE color spaces. It is based on tristimulus values that represent a quantitative measure for all colors. Chromaticity coordinates (x, y, z) are derived from these values and are used to represent colors in a two-dimensional plane \cite{Kang}. Y represents luminance, Z approximates blue in CIE RGB, and X is a nonnegative mix of the three CIE RGB curves. Defining Y as luminance ensures that the XZ plane contains all possible chromaticities at a given brightness.

To convert from the RGB to the CIE XYZ, the following transformation matrix is applied:
\begin{equation}
\begin{bmatrix} X \\ Y \\ Z \end{bmatrix} =
\begin{bmatrix}
0.4124564 & 0.3575761 & 0.1804375 \\
0.2126729 & 0.7151522 & 0.0721750 \\
0.0193339 & 0.1191920 & 0.9503041
\end{bmatrix}
\begin{bmatrix} R' \\ G' \\ B' \end{bmatrix}
\label{eq:rgb_to_xyz}
\end{equation}
where \( R', G', B' \) are the normalized RGB values in the range \([0,1]\), obtained as:
\[
R' = \frac{R}{255}, \quad G' = \frac{G}{255}, \quad B' = \frac{B}{255}
\]

The inverse transformation, converting from CIE XYZ to RGB, is given by:
\begin{equation}
\begin{bmatrix} R' \\ G' \\ B' \end{bmatrix} =
\begin{bmatrix}
  3.2404542 & -1.5371385 & -0.4985314 \\
 -0.9692660 &  1.8760108 &  0.0415560 \\
  0.0556434 & -0.2040259 &  1.0572252
\end{bmatrix}
\begin{bmatrix} X \\ Y \\ Z \end{bmatrix}
\label{eq:xyz_to_rgb}
\end{equation}
The resulting values \( R', G', B' \) are then converted back to the standard RGB range \([0,255]\) as follows:
\[
R = 255 \times R', \quad G = 255 \times G', \quad B = 255 \times B'
\]
If necessary, gamma correction should be applied to ensure accurate color representation.

CIEXYZ is widely used in various applications, including image processing, computer vision, medical imaging, and color calibration, i.e., for tasks requiring high color accuracy.  Recent advancements propose using self-supervised learning (SSL) frameworks to reconstruct CIE-XYZ images from sRGB images, reducing the dependency on extensive paired data and outperforming existing methods \cite{Barzel2024SEL-CIE:}

CIEXYZ has derivatives such as CIELUV and CIELAB, both of which were developed to enhance perceptual uniformity \cite{Pointer2009A}. This means that within these color spaces, equal numerical differences correspond more closely to uniform perceived color variations \cite{Lozhkin2021Investigation}.

\subsubsection*{CIE L*a*b*}
\newtheorem{definition}{Definition}
\begin{definition}
A uniform color space (UCS) refers to a color space in which the same spatial distance between any two colors corresponds to the same amount of perceived color difference.
\end{definition} 
The CIELAB color model is known for its computational complexity but is widely regarded as one of the most perceptually UCS. It is considered a universal model for color difference.  It was introduced by the CIE in 1976 as a refinement of the CIE XYZ model, with the primary objective of quantifying color differences more effectively while also serving as a fundamental framework for color appearance modeling \cite{Fairchild1993}. It represents color via three components: L*, which indicates perceptual lightness, and a* and b*, which correspond to the four distinct colors that humans can perceive: red, green, blue, and yellow.

CIELAB and CIELUV color spaces provide, for the first time, numerical correlates of lightness, hue, chroma, and saturation \cite{Fairchild2013}.

In three-dimensional space, the Lab* model maintains near-uniform contrast, forming an almost isotropic orthogonal system derived from XYZ visual stimuli, with adjustments based on a reference or adaptive illuminant \cite{review_prev}. Although the model follows a curvilinear structure rather than a perfectly uniform contrast distribution, it can be applied using linear transformations. A key advantage of CIELAB is its device-independent nature, allowing seamless conversion between CIE XYZ and other color spaces.

Given its independence from specific hardware, CIELAB plays a crucial role in desktop color management and ensures consistent reproduction across different devices \cite{Hasan2012}. It is effectively utilized in several domains, including food coloring to estimate the types of colorants in products like boiled sausages \cite{Jovanović2017Verification}, image processing for color enhancement while preserving the color tone \cite{Chiang2018Color}, and surface grading \cite{López‐García2005Fast}.

A limitation of CIELAB is its lack of complete chromatic adaptation. It applies the von Kries transform method directly in the XYZ color space rather than first converting to the LMS color space for better accuracy. As a result, CIELAB tends to underperform when employing a non-reference white point, which limits its effectiveness as an adaptation model.

The incorrect transformation also causes its uneven blue hue, which shifts toward purple with changes in L. CIELAB lacks perceptual uniformity, especially regarding blue shades, although it was initially planned as UCS. Highly chromatic blues tend to appear more purple as chroma decreases due to deficiencies in the CIELAB equations and the use of CIE Standard Observers \cite {Moroney2003A}, \cite{Cinko2019Dependence}. CIELAB is also known for deficiencies in perceptual uniformity in the dark regions of color space \cite{Sharma2012The}.

The conversion between the RGB and CIE LAB color spaces is a multi-step process that involves an intermediate transformation to the CIE XYZ space.
To convert from RGB to LAB, the following steps are applied:
\begin{enumerate}
    \item Normalize RGB values to the range \([0,1]\) and apply gamma correction:
    \[
    R' = \begin{cases} 
    \frac{R}{255} / 12.92, & \text{if } \frac{R}{255} \leq 0.04045 \\ 
    \left(\frac{R}{255} + 0.055 \right) / 1.055)^{2.4}, & \text{otherwise}
    \end{cases}
    \]
    \[
    G', B' \text{ follow the same transformation.}
    \]
    \item Convert linearized RGB to XYZ using the transformation matrix provided above in Equation \ref{eq:rgb_to_xyz}.
    \item Normalize XYZ values using the reference white \( (X_n, Y_n, Z_n) \) (D65 standard):
    \[
    X_n = 95.047, \quad Y_n = 100.000, \quad Z_n = 108.883
    \]
    \[
    x = \frac{X}{X_n}, \quad y = \frac{Y}{Y_n}, \quad z = \frac{Z}{Z_n}
    \]
    \item Apply the nonlinear transformation:
    \[
    f(t) = \begin{cases} 
    t^{1/3}, & \text{if } t > 0.008856 \\ 
    \frac{7.787t + \frac{16}{116}}, & \text{otherwise}
    \end{cases}
    \]
    \item Compute LAB components:
    \[
    L = 116 \cdot f(y) - 16
    \]
    \[
    a = 500 \cdot (f(x) - f(y))
    \]
    \[
    b = 200 \cdot (f(y) - f(z))
    \]
\end{enumerate}

The inverse transformation follows these steps:
\begin{enumerate}
    \item Compute the intermediate values:
    \[
    y = \frac{L + 16}{116}, \quad x = \frac{a}{500} + y, \quad z = y - \frac{b}{200}
    \]
    \item Apply the inverse nonlinear transformation:
    \[
    f^{-1}(t) = \begin{cases} 
    t^3, & \text{if } t > 0.206893 \\ 
    \frac{t - \frac{16}{116}}{7.787}, & \text{otherwise}
    \end{cases}
    \]
    \[
    X = X_n f^{-1}(x), \quad Y = Y_n f^{-1}(y), \quad Z = Z_n f^{-1}(z)
    \]
    \item Convert XYZ to linear RGB using Equation \ref{eq:xyz_to_rgb}.
    \item Apply gamma correction:
    \[
    R = \begin{cases} 
    255 \times (1.055 R'^{1/2.4} - 0.055), & \text{if } R' > 0.0031308 \\ 
    255 \times (12.92 R'), & \text{otherwise}
    \end{cases}
    \]
    \[
    G, B \text{ follow the same transformation.}
    \]
\end{enumerate}

\subsubsection*{CIE L*U*V*}
Uniformity is an important concern that needs to be addressed in color models. A uniform color space is a space in which the
change in the color coordinates corresponds to a similar change in the visible colors.  LUV (CIE 1976), one of the CIE-derived color spaces, is recommended as a good approximation of uniform color space \cite{building_cieluv}.

LUV space (where L stands for luminance, whereas U and V represent chromaticity values of color images). CIELUV employs a Judd-type method for white point adaptation, unlike CIELAB, which utilizes a von Kries transformation \cite{Judd1940}. This approach can yield valuable results when dealing with a single light source, but it may also predict nonexistent colors (i.e., those falling outside the spectral locus) when used as a chromatic adaptation transformation \cite{Fairchild2013}. The translational adaptation method applied in CIELUV has been demonstrated to be ineffective at accurately predicting corresponding colors \cite{Alman1989}.

CIE LUV enhances various image processing and analysis tasks due to its ability to accurately represent color differences. Its applications range from image compression and color segmentation to defect detection and remote sensing. By using CIE LUV, image analysis techniques such as segmentation and shadow detection benefit from the space's ability to represent chromatic information effectively, e.g., shadow elimination in traffic flow analysis \cite{Rico-Fernández2019A}.

The conversion between the RGB and CIE LUV color spaces involves an intermediate step through the CIE XYZ color space.
The conversion from RGB to CIE LUV follows these steps:
\begin{enumerate}
    \item Normalize RGB values and apply gamma correction:
    \[
    R' = \begin{cases} 
    \frac{R}{255} / 12.92, & \text{if } \frac{R}{255} \leq 0.04045 \\ 
    \left(\frac{R}{255} + 0.055 \right) / 1.055)^{2.4}, & \text{otherwise}
    \end{cases}
    \]
    \[
    G', B' \text{ follow the same transformation.}
    \]
    \item Convert linearized RGB to XYZ using Equation \ref{eq:rgb_to_xyz}.
    \item Compute the chromaticity coordinates:
    \[
    u' = \frac{4X}{X + 15Y + 3Z}, \quad v' = \frac{9Y}{X + 15Y + 3Z}
    \]
    \item Calculate LUV components:
    \[
    L = \begin{cases} 
    116 \cdot (Y/Y_n)^{1/3} - 16, & \text{if } Y/Y_n > 0.008856 \\ 
    903.3 \cdot (Y/Y_n), & \text{otherwise}
    \end{cases}
    \]
    \[
    u = 13L (u' - u'_n), \quad v = 13L (v' - v'_n)
    \]
    where \( u'_n, v'_n \) are the white point chromaticity coordinates:
    \[
    u'_n = \frac{4X_n}{X_n + 15Y_n + 3Z_n}, \quad v'_n = \frac{9Y_n}{X_n + 15Y_n + 3Z_n}
    \]
    with reference white values for D65:
    \[
    X_n = 95.047, \quad Y_n = 100.000, \quad Z_n = 108.883
    \]
\end{enumerate}

The inverse conversion follows these steps:
\begin{enumerate}
    \item Compute the intermediate values:
    \[
    u' = \frac{u}{13L} + u'_n, \quad v' = \frac{v}{13L} + v'_n
    \]
    \item Calculate Y from L:
    \[
    Y = \begin{cases} 
    \left(\frac{L + 16}{116} \right)^3 Y_n, & \text{if } L > 8 \\ 
    \frac{L}{903.3} Y_n, & \text{otherwise}
    \end{cases}
    \]
    \item Reconstruct X and Z:
    \[
    X = \frac{9Yu'}{4v'}
    \]
    \[
    Z = \frac{Y(12 - 3u' - 20v')}{4v'}
    \]
    \item Convert XYZ to linear RGB using Equation \ref{eq:xyz_to_rgb}.
    \item Apply gamma correction:
    \[
    R = \begin{cases} 
    255 \times (1.055 R'^{1/2.4} - 0.055), & \text{if } R' > 0.0031308 \\ 
    255 \times (12.92 R'), & \text{otherwise}
    \end{cases}
    \]
    \[
    G, B \text{ follow the same transformation.}
    \]
\end{enumerate}

\subsubsection*{CIE color appearance models}
\newtheorem{definitionn}{Definition}
\begin{definition}
A color appearance model (CAM) is a mathematical framework designed to represent the perceptual elements of human color vision \cite{Fairchild2013}.
\end{definition} 

A color model establishes a coordinate system to quantify colors, e.g., RGB, XYZ, and CMYK models.  In contrast, CAM addresses instances where the perceived color does not correspond to the physical measurement of the stimulus source, depending on various viewing conditions. 

Uniform color spaces described above, namely, CIELAB and CIELUV, can be considered as the first step towards color appearance models. These color spaces have straightforward chromatic adaptation transformations and predictors for lightness, chroma, and hue.

In 1997, after developing CIELAB, the CIE aimed to create a comprehensive color appearance model, resulting in CIECAM97s. It has been criticized for its complexity, which has hindered its widespread adoption \cite{Fairchild2001A}.

Next, the refined version of CIECAM97s, CIECAM02, was developed in 2002. It performs better while also being simpler. Besides the basic CIELAB model, CIECAM02 is the closest to an internationally accepted "standard” for a comprehensive CAM.  CIECAM02 incorporates a linear chromatic adaptation transform, a non-linear response compression function, and adjustments for perceptual attribute calculations \cite{Moroney2002The}. It outperforms CIECAM97s, especially in predicting saturation, making it a potential replacement for image applications \cite{Li2002The}.

RLAB aims to address the major shortcomings of CIELAB, particularly in image reproduction. In contrast to CIELAB, RLAB implements an appropriate von Kries adjustment \cite{Fairchild1993}. It is not suitable for other applications, which limits its usage.


\subsubsection*{CIE Color Difference Methods}

CIE Color Difference Methods quantify perceived color differences, which is essential in industries like textiles, digital imaging, and color reproduction. The CIE has developed several color difference formulas over the years, including CIE76 \cite{Robertson1977The}, CIE94 \cite{mcdonald1995cie94}, and CIE2000 \cite{luo2001development} methods, which are very important in modern image processing \cite{Proskuriakov2021Features}.

CIE76 has the highest Delta E values across various image formats, irrespective of the dominant color. In contrast, CIE94 and CIE2000 show minor deviations and substantially differ from CIE76, particularly during maximum image reduction. These methods preserve the main Delta E peaks, indicating reliability in color difference calculations. The choice between these formulas often depends on factors such as industry familiarity and common practice \cite{Robertson1977The}.

The CIE76 color difference formula (\(\Delta E_{76}\)) is the simplest and is based on the Euclidean distance in the CIELAB color space \cite{Robertson1977The}:
\begin{equation}
\Delta E_{76} = \sqrt{(L_2 - L_1)^2 + (a_2 - a_1)^2 + (b_2 - b_1)^2}
\label{eq:deltaE76}
\end{equation}
where \(L\), \(a\), and \(b\) are the CIELAB color space coordinates of two colors.

The CIE94 color difference formula (\(\Delta E_{94}\)) introduces weighting functions to improve perceptual uniformity \cite{mcdonald1995cie94}:
\begin{equation}
\Delta E_{94} = \sqrt{\left(\frac{\Delta L}{k_L S_L}\right)^2 + \left(\frac{\Delta C}{k_C S_C}\right)^2 + \left(\frac{\Delta H}{k_H S_H}\right)^2}
\label{eq:deltaE94}
\end{equation}
where \(\Delta L = L_2 - L_1\) is the lightness difference, \(\Delta C = C_2 - C_1\) is the chroma difference in CIELAB,
 \(\Delta H = \sqrt{(a_2 - a_1)^2 + (b_2 - b_1)^2 - \Delta C^2}\) is the hue difference,
 \(S_L, S_C, S_H\) are scaling functions for lightness, chroma, and hue,
 \(k_L, k_C, k_H\) are weighting factors (typically set to 1 for standard conditions).

The CIEDE2000 color difference formula (\(\Delta E_{00}\)) improves upon previous models by incorporating corrections for hue rotation, chroma scaling, and lightness differences. It is given by \cite{luo2001development}:

\begin{strip}
\begin{equation}
\Delta E_{00} =
\sqrt{%
    \left(\frac{\Delta L'}{k_L S_L}\right)^2 +
    \left(\frac{\Delta C'}{k_C S_C}\right)^2 +
    \left(\frac{\Delta H'}{k_H S_H}\right)^2 +
    R_T
    \left(\frac{\Delta C'}{k_C S_C}\right)
    \left(\frac{\Delta H'}{k_H S_H}\right)
}
\label{eq:deltaE00}
\end{equation}
\end{strip}

where \(\Delta L'\) is the lightness difference,
\(\Delta C'\) is the chroma difference,
\(\Delta H'\) is the hue difference,
 \(S_L\), \(S_C\), and \(S_H\) are the weighting functions for lightness, chroma, and hue,
\(k_L, k_C, k_H\) are parametric weighting factors (typically set to 1),
\(R_T\) is a rotation term accounting for chroma and hue interactions.

For simple and fast calculations, CIE76 is the easiest to compute but lacks perceptual accuracy. CIE94 improves perceptual consistency with weighting factors, making it better suited for industrial applications. CIEDE2000 is the most perceptually uniform for the highest accuracy and is recommended for modern applications.

\begin{figure*}
     \centering
     \begin{subfigure}[b]{0.3\textwidth}
         \centering
        \includegraphics[width=\linewidth]{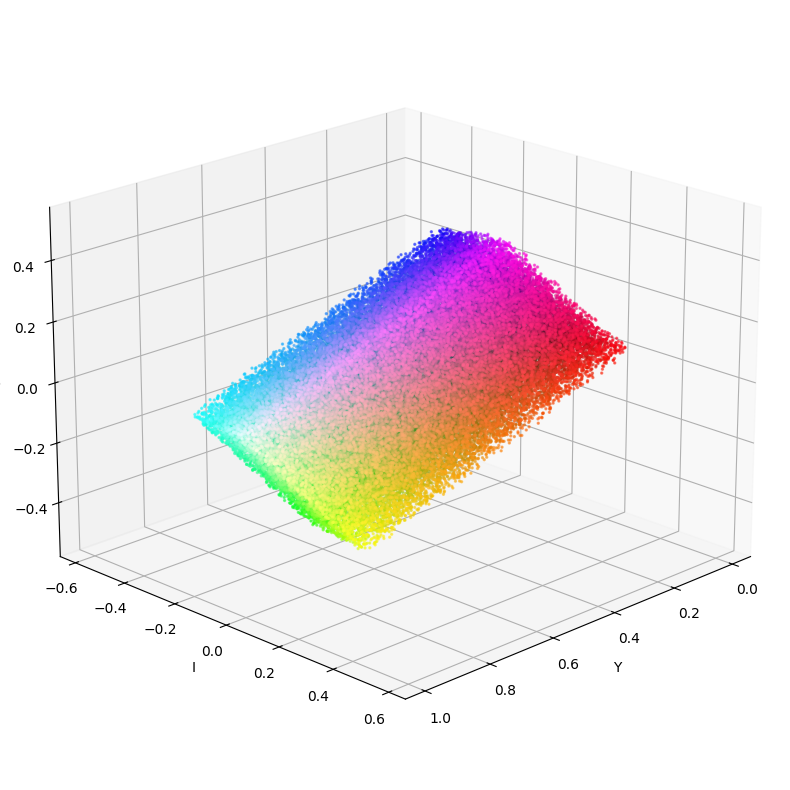}
        \caption{RGB gamut visualized in YIQ}
        \label{fig:yiq}
     \end{subfigure}
     \hfill
     \begin{subfigure}[b]{0.65\textwidth}
        \centering
        \includegraphics[width=\linewidth]{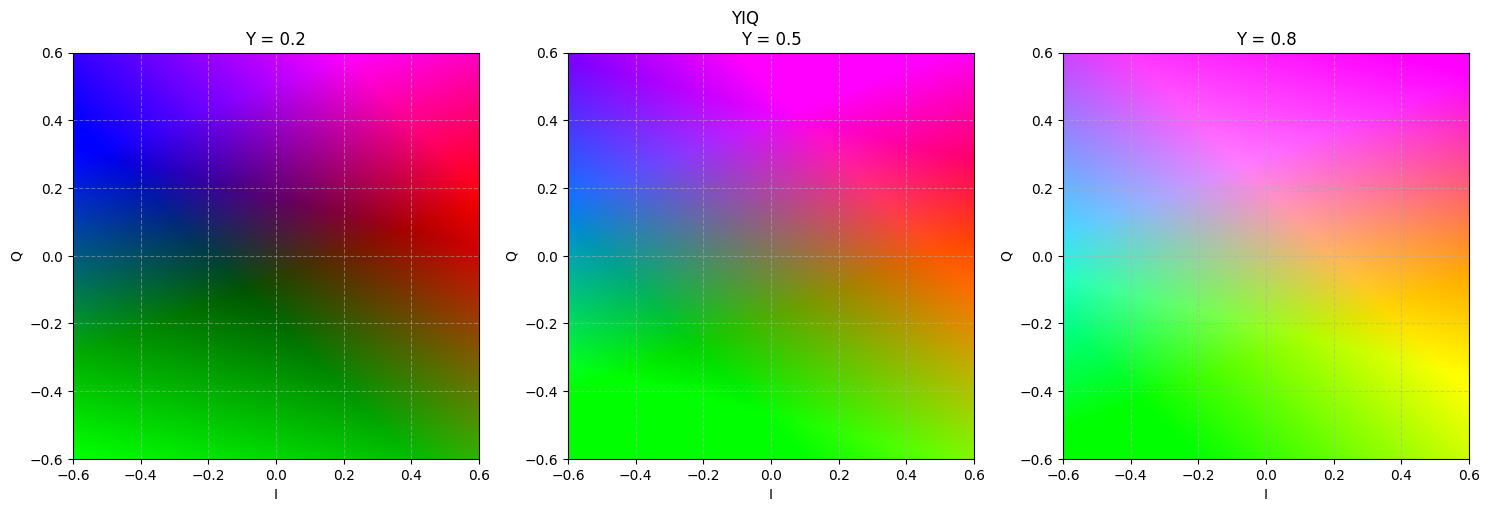}
        \caption{Cross-sections of the YIQ color space at constant luminance values: Y = 0.2, Y = 0.5, and Y = 0.8. The plots illustrate the variation of the chromatic distribution in the I–Q plane as luminance increases.}
        \label{fig:yiq_slice}
     \end{subfigure}
        \caption{Luma-Chroma models visual representation (YIQ)}
        \label{fig:y_components}
\end{figure*}

\subsection*{Luma-Chroma Models}
The YIQ, YCbCr, and YUV color models are commonly grouped as luma-chroma models, as they separate an image's luminance (luma) and chrominance (chroma) components. These models are fundamental in video and digital imaging applications, where maintaining perceptual quality while minimizing data requirements is crucial.
As shown in Figure \ref{fig:y_components}, the same RGB cube takes different forms in alternative representations. Specifically, Figure \ref{fig:yiq} demonstrates the projection into the YIQ system, where luminance and opponent channels (I, Q) are separated. 

\subsubsection*{YIQ}
The YIQ color model is a widely recognized color space primarily used in television broadcasting. Originally introduced by the National Television Standards Committee (NTSC) in 1953, the model was designed to maximize perceptual resolution while maintaining compatibility with black-and-white television signals \cite{Schwarz1987}.

The YIQ color model consists of three components:
\begin{itemize}
    \item \textbf{Y (Luminance)}: Represents the brightness of the image and is derived from a weighted sum of the red (R), green (G), and blue (B) components.
    \item \textbf{I (In-phase)}: Encodes chrominance information along a blue-green to orange vector.
    \item \textbf{Q (Quadrature)}: Encodes chrominance information along a yellow-green to magenta vector.
\end{itemize}

As shown in Figure \ref{fig:yiq_slice}, the shape and extent of the gamut in the I–Q plane depend on the luminance value: lower Y results in a more compressed distribution, while higher Y expands the chromatic range along both opponent dimensions.

The YIQ color space was developed to efficiently encode color information in television broadcasts. It was optimized for the human visual system by allocating most of the bandwidth to luminance (Y) while limiting the bandwidth for chrominance (I and Q) since the human eye is less sensitive to color variations than to brightness differences \cite{Schwarz1987}. This separation allows for efficient compression and transmission, ensuring backward compatibility with black-and-white televisions. Additionally, the YIQ color space is device-dependent, meaning the colors displayed depend on the specific monitor settings and phosphors used \cite{Khalili2013}. The transformation equations between RGB and YIQ are given below.

\begin{equation}
\begin{bmatrix}
Y \\
I \\
Q
\end{bmatrix}
=
\begin{bmatrix}
0.299 & 0.587 & 0.114 \\
0.596 & -0.274 & -0.322 \\
0.211 & -0.523 & 0.312
\end{bmatrix}
\begin{bmatrix}
R \\
G \\
B
\end{bmatrix}
\end{equation}

\begin{equation}
\begin{bmatrix}
R \\
G \\
B
\end{bmatrix}
=
\begin{bmatrix}
1.000 & 0.956 & 0.621 \\
1.000 & -0.272 & -0.647 \\
1.000 & -1.106 & 1.703
\end{bmatrix}
\begin{bmatrix}
Y \\
I \\
Q
\end{bmatrix}
\end{equation}

These equations are based on the model used in NTSC television broadcasting \cite{poynton2012digital}.

Although originally developed for analog television, the principles behind YIQ remain relevant in modern image processing applications. By embedding chrominance data in the wavelet domain of the luminance component, compression algorithms using YIQ achieve better performance compared to traditional methods like JPEG \cite{Campisi2002}. Additionally, YIQ proves to be effective in digital watermarking, where the ability to separate luminance and chrominance allows watermarks to be embedded in perceptually less significant areas, enhancing robustness against compression and noise attacks. Discrete wavelet transform-based watermarking algorithms in the YIQ space have shown improved security and resilience \cite{Khalili2013, Cui2018}.

Furthermore, YIQ is utilized in image recognition, such as in robotic tomato harvesting, where the I-component enhances color feature extraction, improving segmentation accuracy and robustness against lighting variations \cite{Zhao2016}. YIQ also contributes to color image enhancement by adjusting the luminance (Y) component while preserving chrominance, which results in better lightness, contrast, and image quality compared to traditional methods. Fuzzy logic-based enhancement techniques have shown superior results in terms of entropy, mean square error, and natural image quality metrics \cite{Daway2020}.

\subsubsection*{YUV}
The YUV color model is a widely used color space primarily designed for image and video processing applications, especially in broadcasting and compression tasks. It was introduced as part of the NTSC color television standard in the 1950s and later adapted in other systems, such as PAL and SECAM \cite{Cardone2023}.

The YUV model divides the color information into three components:
\begin{itemize}
    \item \textbf{Y (Luminance)}: Represents the brightness of the image and is calculated as a weighted sum of the red (R), green (G), and blue (B) components. It is crucial in preserving the intensity or luminance of the image.
    \item \textbf{U (Chrominance Blue)}: Encodes the chromatic information by capturing the blue differences.
    \item \textbf{V (Chrominance Red)}: Encodes the chromatic information by capturing the red differences.
\end{itemize}

The YUV color model is widely used in broadcasting and video applications because it efficiently separates luminance (Y) from chrominance (U and V). This separation is important since the human eye is more sensitive to brightness variations than to color changes, making the Y channel more crucial for perceived image quality. This characteristic enables better compression without perceptible loss of quality, as the chrominance channels (U and V) can be compressed more aggressively than the luminance channel. This advantage is particularly important in the context of analog television and modern digital image compression methods \cite{Domino2022, Cardone2023}. The transformation equations between RGB and YUV are given below \cite{gonzalez2018digital}.

\begin{equation}
\begin{bmatrix}
Y \\
U \\
V
\end{bmatrix}
=
\begin{bmatrix}
0.299 & 0.587 & 0.114 \\
-0.147 & -0.289 & 0.436 \\
0.615 & -0.515 & -0.100
\end{bmatrix}
\begin{bmatrix}
R \\
G \\
B
\end{bmatrix}
\end{equation}

\begin{equation}
\begin{bmatrix}
R \\
G \\
B
\end{bmatrix}
=
\begin{bmatrix}
1.000 & 0.000 & 1.140 \\
1.000 & -0.396 & -0.581 \\
1.000 & 2.029 & 0.000
\end{bmatrix}
\begin{bmatrix}
Y \\
U \\
V
\end{bmatrix}
\end{equation}

The YUV color model is advantageous in various applications. For example, it is frequently used in image compression, where it allows for more efficient data storage by compressing the U and V channels more than the Y channel. One of the main benefits of YUV in image compression is its alignment with human visual perception, where the majority of the image information perceived by the human eye is contained in the Y channel. This separation allows for a reduction in the computational load while preserving the visual quality of the image \cite{Cardone2023}.

In video watermarking and digital security, YUV's ability to separate luminance from chrominance makes it effective for embedding watermarks in less perceptually significant areas of an image. This enhances robustness against attacks like compression and noise \cite{Khalili2013, Koju2015}. Furthermore, YUV has been applied in face super-resolution techniques, where the luminance-chrominance separation helps preserve the identity of facial features while enhancing image quality \cite{Kim2021}.

In terms of agriculture, the YUV color space has been utilized to classify the ripeness of palm oil fruit, demonstrating its effectiveness in automatic classification tasks where rapid and reliable color feature extraction is required to detect the ripeness of agricultural products \cite{Sabri2018}.

Moreover, the YUV color space is beneficial for applications that require skin detection or segmentation. In these areas, the model helps improve the accuracy of skin region detection, especially under challenging lighting conditions. The separation of luminance from chrominance channels makes it more robust against lighting variations, making it a reliable choice for tasks such as face recognition and human-computer interaction \cite{Zaher2014}. The ability to distinguish between the intensity and chromatic components is also advantageous in cloud removal from satellite imagery, where the Y channel can be processed independently of the chrominance channels to remove thin clouds more effectively \cite{Wen2021}.

\subsubsection*{YCbCr}
The YCbCr color model is a digital color space widely used in image and video processing, particularly in compression and broadcast applications. It is derived from the YUV color model, with the key difference being that YUV is an analog system, while YCbCr is a digital system. This model was developed as a part of the digital video and television industry to efficiently represent color information by separating luminance from chrominance, thus optimizing data transmission and storage. YCbCr is not an absolute color model but a scaled and offset version of YUV, specifically designed for digital processing, offering greater compatibility with modern digital formats \cite{Koju2015, Khalili2013}.

The YCbCr model is made up of three components:
\begin{itemize}
    \item \textbf{Y (Luminance)}: Represents the brightness of the image, capturing the intensity of light, which is a weighted sum of the RGB components.
    \item \textbf{Cb (Chrominance Blue)}: Encodes the difference between the blue color channel and the luminance component.
    \item \textbf{Cr (Chrominance Red)}: Encodes the difference between the red color channel and the luminance component.
\end{itemize}

This separation of luminance (Y) from chrominance (Cb and Cr) is particularly effective for compressing video signals because human vision is more sensitive to variations in brightness than to color differences. By allocating more bandwidth to the luminance component, YCbCr ensures efficient use of data while still representing the full spectrum of colors.

The YCbCr model was initially adopted for television broadcasting, as it was more compatible with black-and-white television sets than models like RGB. It has since become a key component in various image and video compression standards, including JPEG, MPEG, and H.264. This model remains relevant today, especially in modern compression algorithms, where the ability to separate the luminance and chrominance channels allows for greater compression efficiency without significant loss of perceptual quality \cite{Koju2015, Khalili2013}. The transformation equations between RGB and YCbCr are given below.

\begin{equation}
\begin{bmatrix}
Y \\
Cb \\
Cr
\end{bmatrix}
=
\begin{bmatrix}
0.299 & 0.587 & 0.114 \\
-0.1687 & -0.3313 & 0.5000 \\
0.5000 & -0.4187 & -0.0813
\end{bmatrix}
\begin{bmatrix}
R \\
G \\
B
\end{bmatrix}
+
\begin{bmatrix}
0 \\
128 \\
128
\end{bmatrix}
\end{equation}

\begin{equation}
\begin{bmatrix}
R \\
G \\
B
\end{bmatrix}
=
\begin{bmatrix}
1.000 & 0.000 & 1.402 \\
1.000 & -0.344 & -0.714 \\
1.000 & 1.772 & 0.000
\end{bmatrix}
\begin{bmatrix}
Y - 16 \\
Cb - 128 \\
Cr - 128
\end{bmatrix}
\end{equation}

These equations comply with the ITU-R BT.601 standard, which is widely used in digital television and video processing \cite{winkler2005digital}.

In practical applications, YCbCr has shown its effectiveness across a range of domains. In image watermarking, for instance, YCbCr's separation of luminance and chrominance makes it possible to embed watermarks in areas of the image that are less perceptible to the human eye, thereby improving the robustness of the watermark against common attacks like compression or noise \cite{Koju2015, Khalili2013}. Furthermore, the YCbCr model is widely used in skin color detection for human-computer interaction systems, as it allows for better segmentation of skin regions from the background, crucial for applications such as gesture recognition \cite{Islam2020}.

In agricultural image analysis, YCbCr has been used to classify the ripeness of palm oil fresh fruit bunches (FFB). Research has shown that the YCbCr model, along with other color models, helps accurately classify various ripeness stages of agricultural products based on their color characteristics, proving particularly useful in automated harvesting systems \cite{Sabri2018}. Additionally, YCbCr has been applied in disease detection on potato leaves, where it demonstrated superior performance compared to other color spaces in detecting diseased areas in complex, real-world environments \cite{Johnson2021}.

\subsection*{LMS color space}

\begin{figure}
    \centering
    \includegraphics[width=\linewidth]{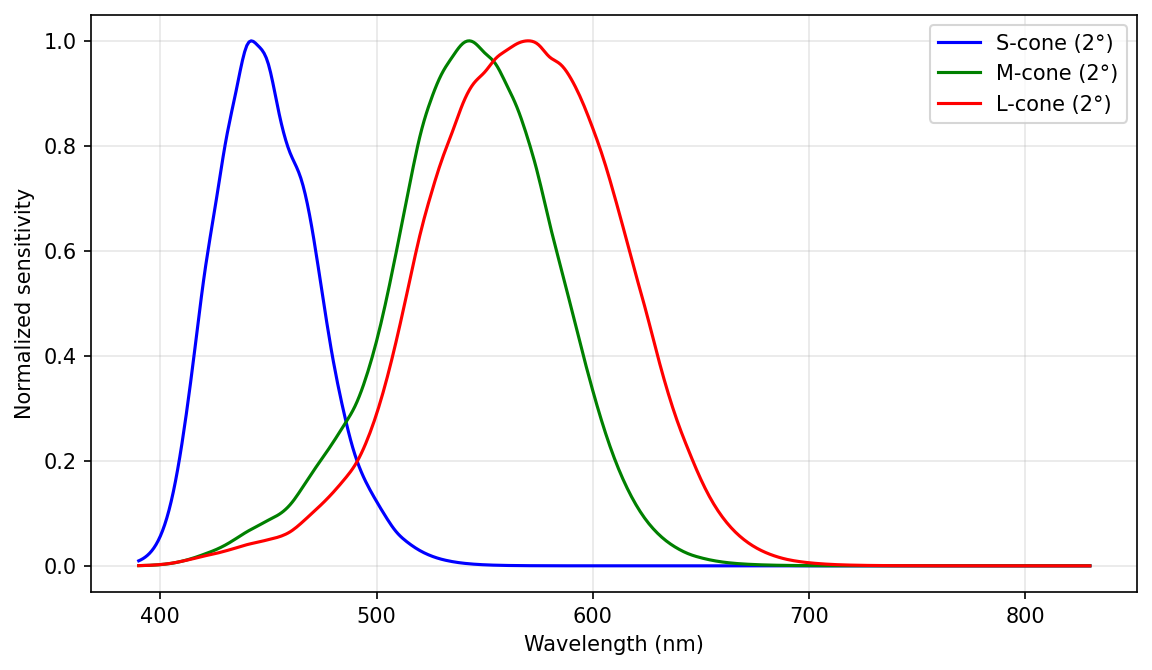}
    \caption{Cone spectral sensitivity functions (Stockman \& Sharpe, 2000) for the long- (L), medium- (M), and short- (S) wavelength sensitive cones, plotted for the 2° standard observer. Each function is normalized to its maximum response to highlight relative differences in peak sensitivity and spectral bandwidth.}
    \label{fig:lms}
\end{figure}

The LMS (Long-Medium-Short) color space is a model that represents colors based on the human visual system, specifically the sensitivity of the three types of cone cells in the retina. Each cone type responds to a specific range of wavelengths: L (Long) cones are most sensitive to longer wavelengths in the red spectrum, M (Medium) cones respond to mid-range wavelengths associated with green, and S (Short) cones are primarily sensitive to shorter wavelengths in the blue region. This color space plays a fundamental role in color science as it closely mirrors the way humans perceive and distinguish colors \cite{Johnson2010DerivationOA, Romney2009}. Fig. \ref{fig:lms} illustrates these cone fundamentals based on the 2° standard observer data reported by Stockman and Sharpe (2000), which are widely used in color vision research and colorimetry.

One of the key applications of the LMS color space is in color vision deficiency (CVD) simulation and correction. Since color blindness results from anomalies in the cone cells, transforming images into the LMS space allows for precise modifications that enhance color distinction for individuals with Protanopia, Deuteranopia, and Tritanopia (Color Blindness Types)\cite{7155920}. Researchers have implemented LMS-based Daltonization algorithms, which selectively adjust colors within the LMS color space to compensate for lost color perception while preserving brightness and contrast. These approaches have proven particularly useful for traffic signals, navigation systems, and educational tools, helping colorblind individuals distinguish between colors they would otherwise confuse\cite{Elrefaei_2018}. This ability to retain subtle color distinctions makes LMS valuable in medical imaging, enhancing contrast in retinal scans, pathology slides, and histology images for better diagnosis. LMS color space is not only used for human vision modeling but also in areas like adaptive filtering, sonar processing, and halftone image enhancement. In halftone printing, LMS helps improve edge sharpness and reduces the dot gain effect, leading to better print quality \cite{4202590}. Despite these broader applications, the primary significance of LMS color space lies in image processing techniques for color vision deficiency and perceptual color transformations, making it a vital tool for color adaptation, accessibility, and improvements in medical imaging.

LMS color space is particularly valuable in research contexts involving color vision deficiencies, as it provides insights into how individuals with various forms of color blindness perceive color. However, since our experiments were designed for participants without color blindness, LMS was deemed less relevant for the objectives of our research.

\subsection*{ICtCp}
ICtCp is a way to show colors for high dynamic range (HDR) and wide color gamut signals (WCG) created by Dolby Laboratories \cite{DolbyICTCP}. It uses three parts to describe color:
\begin{itemize}
    \item I – Intensity
    \item Ct – Blue-yellow chrominance component (named after tritanopia)
    \item Cp – Red-green chrominance component (named after protanopia)
\end{itemize}
The conversion of ICtCp to RGB have been specified by the
ITU-R in Recommendation ITU-R BT.2100-0 (07/2016) \cite{ITUBT2100}, the process goes like this: RGB → LMS → ICtCp. ICtCp is connected to the LMS color space, and to change ICtCp into RGB, first the conversion to LMS must be done, and then the non-linearity functions of the perceptual quantizer (PQ) or hybrid log-gamma (HLG) must be applied. Most commonly, ICtCp is associated with the PQ function.

We do not include the ICtCp color model in our experiments for a few reasons. Firstly, ICtCp it is not a general-purpose color model and is designed specifically for HDR (High Dynamic Range) color representation in the scope of modern video standards. In addition, ICtCp requires matrix operations with PQ/HLG transforms, careful attention to reference white, color primaries, and transfer functions, and is ill-defined without assuming the bit depth, HDR format, and electro-optical transfer function (EOTF). This adds technical overhead for an experiment unless it is specifically about HDR color representation.

\subsection*{Fuzzy Color Models} 
\subsubsection*{Fuzzy Sets}
The concept of fuzzy sets was first introduced by Zadeh \cite{ZADEH1965338}. The main idea behind fuzzy sets is that an element can have a degree of membership anywhere within the interval. In classical (or crisp) sets, which we assume usually, an element can only take membership of 0 or 1. In fuzzy set $A$, the membership function denoted by $\mu_A$, represents membership degree $\mu_A(x)$ of element $x$ to the fuzzy set $A$.

To model uncertainty in fuzzy sets different types of membership functions are used, such as triangular, trapezoidal, sigmoid, and Gaussian membership functions.

Further, defuzzification is applied, a process of converting a fuzzy set to a crisp value. The most widely used defuzzification method is to get the center of gravity of the area under the membership function curve.



\subsubsection*{Fuzzy Color Space}
Color in images is uncertain and imprecise \cite{Shamoi2019, Shamoi2014FuzzyCS}. Nearby pixels can differ from each other, even considering their similarity in human eyes. To deal with the uncertainty of color in images, fuzzy logic has been implemented in image processing due to its ability to deal with uncertain soft boundaries of the crisp color introducing a fuzzy color \cite{sotohidalgo2010}.

\newtheorem{definition3}{Definition}
\begin{definition}A fuzzy color $\boldsymbol{\tilde{C}}$ is a linguistic label whose semantics is defined in a generic color space ${XYZ}$ by a normalized fuzzy subset of $D_X \times D_Y  \times D_Z$ \cite{sotohidalgo2010}. 
\end{definition} For each color fuzzy color $\boldsymbol{\tilde{C}}$ there exists at least one crisp color whose membership to $\boldsymbol{\tilde{C}}$ is 1.

Assuming a fixed color space ${XYZ}$ with $D_x, D_Y$, and $D_Z$ being the domains of the corresponding color component, the extension ofthe  fuzzy color concept, fuzzy color space occurs. 
\begin{definition}
A fuzzy color space $\widetilde{XYZ}$ is a set of fuzzy colors that define a partition $D_x \times D_Y  \times D_Z$.
\end{definition}
 
\subsubsection*{Fuzzy HS*}

Fuzzification of the HS* color spaces is a widely used approach to align with human perception of color. The advantage of the HS* family lies in its perceptual intuitiveness compared to other color spaces \cite{Shamoi2019, Shamir, Amante, Jesus2007, MuragulColorEmotion}.

The fuzzification of HS* color models is performed by defining membership functions for each color component (H, S, and I/V/L). Common choices for these functions include triangular and trapezoidal membership functions. The membership function \(\mu_H\) assigns a membership value to indicate the degree to which a color \(C\) belongs to multiple hues. For example, if the hue is \(15^\circ\), the color may have a membership of \(0.8\) for red and \(0.2\) for orange. A similar approach is applied to the remaining two components.

Due to the semantics of the HS* color space components (Hue, Saturation, and Intensity/Value/Brightness or Lightness), which are commonly used by humans when describing colors, it becomes easier to define meaningful linguistic labels for them \cite{Jesus2007}. The fuzzification of the HS* family is primarily achieved by applying fuzzy logic to the color component values within their respective domains and ranges.

Another reason for the utilization of the HS* family, specifically HSL, is its capability to support high-dimensional derivations and accurately represent tone modifiers in membership functions \cite{sugano_colorspace}.

The advantages of fuzzy HS* color spaces stem from the ability of HS* color models to support linguistic descriptions of color that are closely aligned with human perception. Fuzzy sets further enhance this capability by introducing flexible and non-uniform boundaries. As a result, fuzzy HS* color spaces have a wide range of applications, including image retrieval due to their human-like color descriptions \cite{Jesus2007, younes2007image}, color naming based on linguistic descriptions rather than numerical values, as in RGB \cite{sugano_colorspace}, segmentation \cite{Shamir, Amante}, medical image processing, and design and aesthetics, where linguistic queries can be utilized \cite{Shamoi2016}.


\begin{figure}[t]
    \centering
    \includegraphics[width=\linewidth]{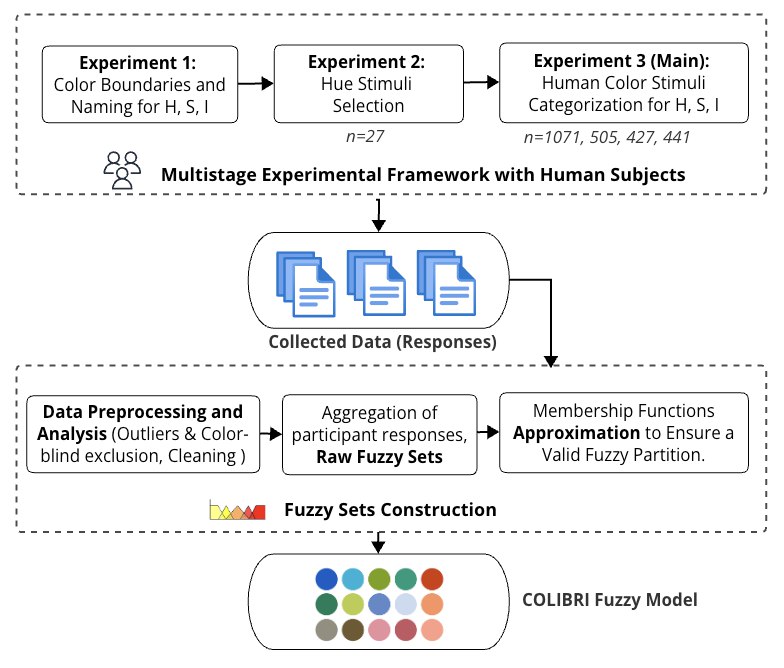}
    \caption{Construction of COLIBRI color model \cite{shamoi2025colibri}}
    \label{fig:placeholder}
\end{figure}

\subsubsection*{COLIBRI}
The Human Perception-Based Fuzzy Color Model - COLIBRI (Color Linguistic-Based Representation and Interpretation) is a novel model designed to bridge the gap between computational color representations and human visual perception. Unlike  traditional color models such as RGB, HSV, CIE Lab, COLIBRI integrates fuzzy set theory and linguistic color categories to account for the inherent uncertainty and subjectivity of human color perception \cite{shamoi2025colibri}.
COLIBRI is based on the observation that humans describe colors linguistically, often with fuzzy boundaries and overlapping categories. Traditional models assign strict numerical values and fixed partitions, which fail to capture these perceptual nuances.  COLIBRI addresses this by constructing fuzzy partitions for hue, saturation, and intensity, enabling colors to belong to multiple categories simultaneously with varying membership degrees.

The construction of COLIBRI followed a three-phase experimental framework (Fig. \ref{fig:placeholder}):
\begin{enumerate}
    \item Perceptual color boundaries and naming - Participants defined linguistic categories and boundaries for hue, saturationa, and intensity.
    \item Hue stimuli selection - Participants refine hue ranges through surveys, indentifying perceptually distingushable steps.
    \item Human Color Stimuli Categorization (Large-scale categorization experiment) - conducted with over 1000 participants (n = 1071 for hue, n = 2496 total including saturation and intensity tasks). The results of experiment was used to create fuzzy membership functions for each color category.
\end{enumerate}

\begin{figure*}[t]
    \centering
    \includegraphics[width=0.7\linewidth]{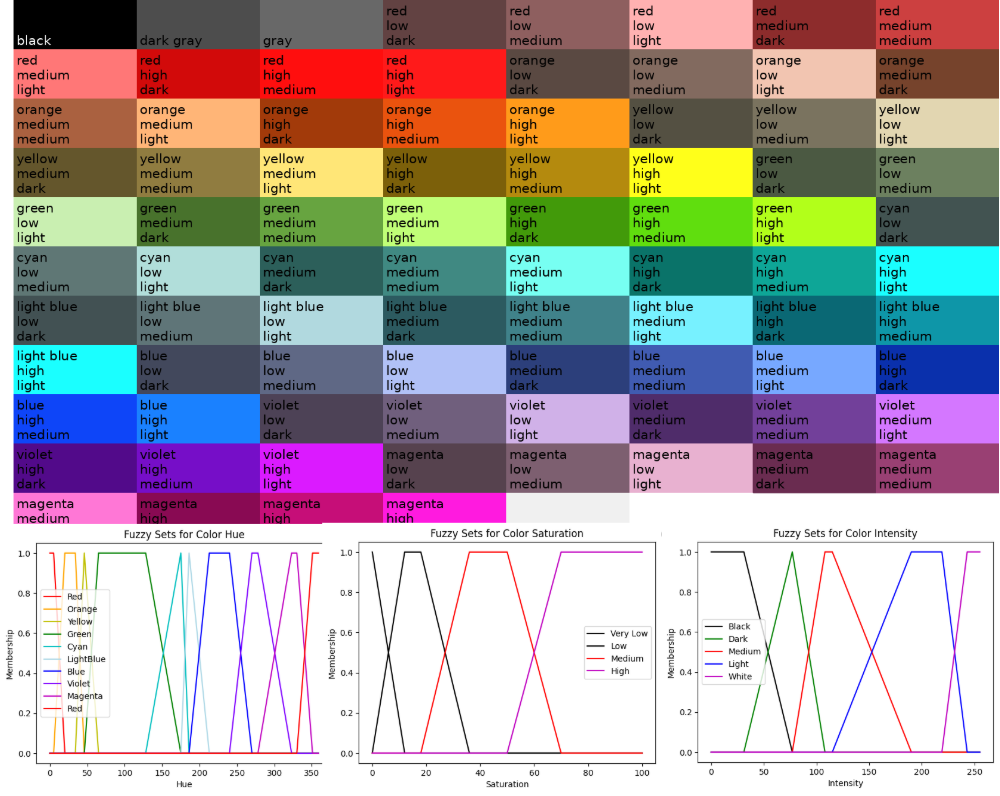}
    \caption{Illustration of COLIBRI color model colors\cite{shamoi2025colibri}
    }
    \label{fig:colibri2}
\end{figure*}

The obstained fuzzy colors are illustrated in Fig. \ref{fig:colibri2}. 

\subsubsection*{Fuzzy CIE}
CIE-based fuzzy color spaces offer an alternative approach to improving automatic color categorization. A key generalization of the CIE model, CIE L*a*b*, is widely used due to its quasi-perceptually homogeneous nature, which ensures that the Euclidean distance between color pairs correlates with human perception of color dissimilarity. When combined with fuzzy logic, this model allows for a more gradual transition in color perception \cite{Benavente2004, Benavente:08, Lammens1995}.

Benavente et al. \cite{Benavente2004} proposed a parametric fuzzy color model using sigmoid and Gaussian functions trained on Berlin and Kay’s color-naming data. This model applies chromaticity-based membership functions across brightness levels to capture variations in color categorization. In a later study, Benavente et al. \cite{Benavente:08} extended their model by incorporating triple sigmoid functions, refining membership estimations across different luminance levels based on psychophysical data. This enhancement further improved automatic color categorization by accounting for complex perceptual transitions in human color vision.

Seaborn et al. \cite{kim2024fuzzy} refined the fuzzy CIE L*a*b* model by implementing fuzzy k-means clustering, enabling adaptive tuning of category memberships for greater robustness across varying lighting conditions. Lammens \cite{Lammens1995} introduced a Gaussian-based fuzzy model that successfully predicted color-naming judgments using graded memberships, effectively handling perceptual ambiguity.

Although the proposed approaches are not limited to CIE L*a*b*, the results of the aforementioned studies demonstrate the advantage of fuzzifying CIE L*a*b* for color-naming tasks.

\subsubsection*{Fuzzy and Munsell}

Though the Munsell color system does not have a linear numerical representation, its standardized color naming system can be widely used as a reference for color naming and linguistic variables in combination with other color models and fuzzy logic.

Tokumaru et al. \cite{TokumaruColorHarmony} apply fuzzy logic to color harmony and design, proposing a system that helps select harmonious color combinations that match the aesthetic preferences of the user. Their system defines fuzzy sets for different types of hue and tone distribution based on Matsuda's Color Coordination, evaluating harmony through fuzzy rules. The Munsell color system provides the "vocabulary" of colors that the fuzzy logic system uses to reason about harmony and aesthetic qualities.

In soil classification, the Munsell color system is used to match the color of the soil. Pegalajar et al. \cite{PEGALAJAR202038} proposed the approach of estimating Munsell color components through values of different color models, like RGB, CIE LAB, and CIE XYZ. Estimated Munsell HVC values are applied with fuzzy rules for each component, creating 238 fuzzy rules with corresponding membership functions.

\section*{Results}
\subsection*{Experiment 1 - Computational Complexity}
In this study, we experimented to analyze the bidirectional conversion of colors between the RGB color space and various alternative color models. Specifically, we examined the following color models: CMY, CMYK, CIE Lab, CIE XYZ, CIE LUV, HSL, HSV, HSI, YIQ, YUV, and YCbCr. The primary objective was to evaluate these conversions' accuracy and computational efficiency using predefined mathematical transformation formulas.

The transformations were implemented using Google Colab. Each input color was represented in the RGB format and subsequently converted into each of the aforementioned color spaces using established conversion equations. Following this, the resulting values in each respective color space were transformed back into the RGB model using the corresponding inverse formulas. 

\begin{figure*}[t]
     \centering
     \begin{subfigure}[b]{0.45\textwidth}
         \centering
         \includegraphics[width=\linewidth]{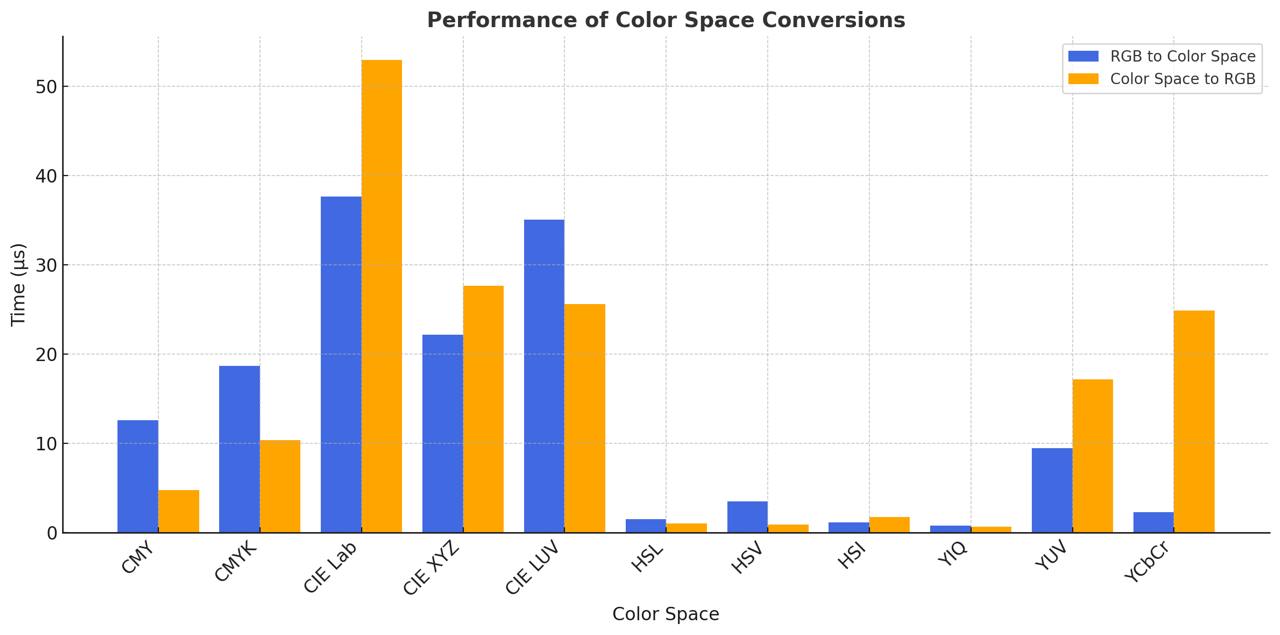}
         \caption{}
         \label{fig:experiment1}
     \end{subfigure}
     \hfill
     \begin{subfigure}[b]{0.45\textwidth}
        \centering
        \includegraphics[width=\linewidth]{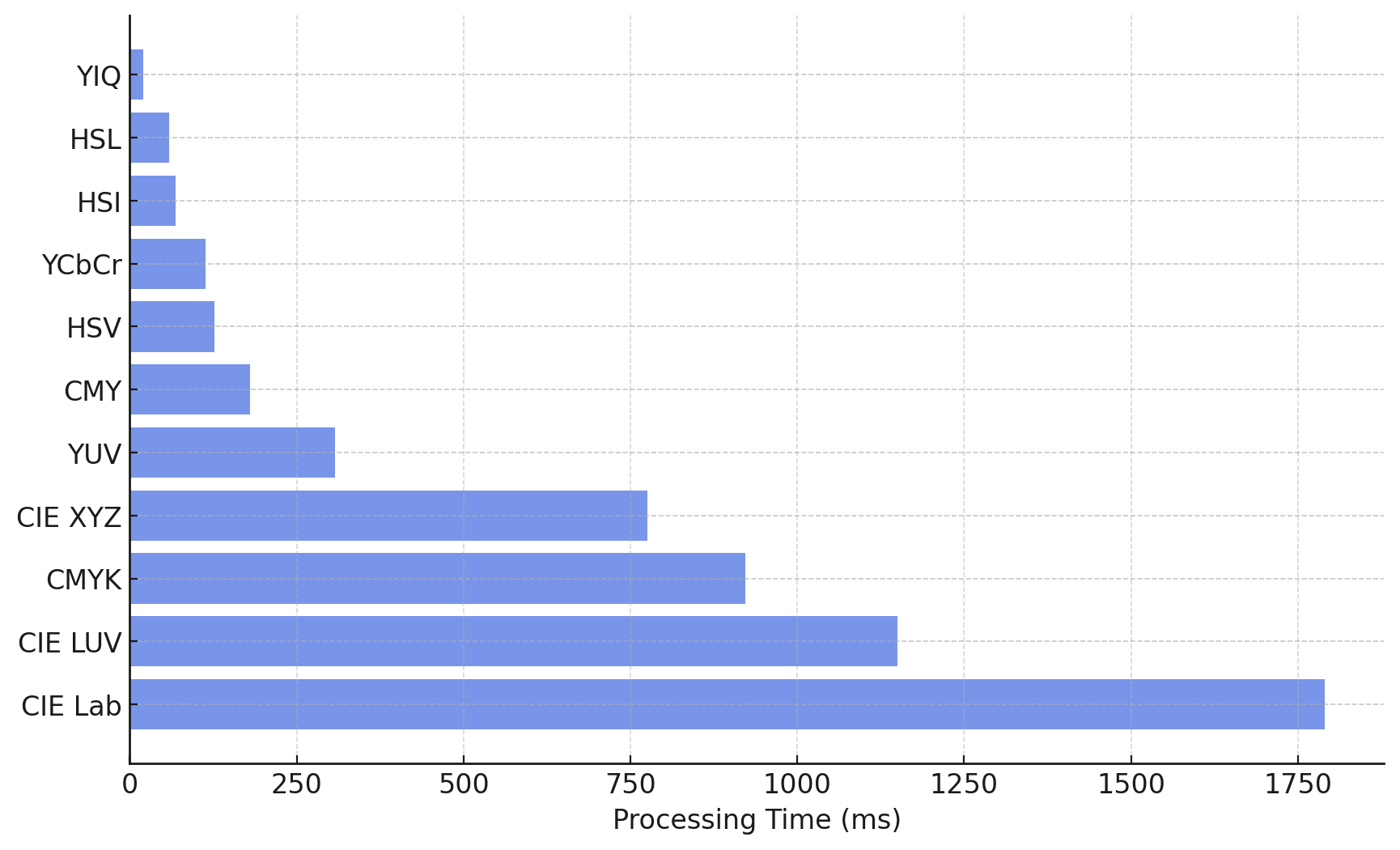}
        \caption{}
        \label{fig:experiment1image}
     \end{subfigure}
        \caption{Experiment 1 results: a) Performance of color models conversions; b) Processing Time for a 200x200 Image in Different Color Spaces}
\end{figure*}

To assess the computational efficiency, we measured the mean execution time for each transformation process. Since function execution time varies due to system workload, network latency, and hardware capabilities, we averaged multiple executions to obtain a representative performance metric. The obtained results are summarized in the Fig. \ref{fig:experiment1}. Mean execution time (mean ± standard deviation) was measured over 7 independent runs, each consisting of 100,000 iterations.

As we can observe from the charts, the fastest transformation in both directions was for the YIQ model, likely due to its simple linear transformation equations. The HSL, HSV, and HSI models also demonstrated very low execution times, as they primarily rely on straightforward arithmetic operations. The CIE Lab and CIE LUV models exhibited the slowest conversions, particularly when converting back to RGB, which can be attributed to the complexity of their non-linear transformations. The CMYK model had a noticeably higher conversion time than the CMY model, as it requires an additional black (K) component computation.

In this experiment, we further converted a 200×200 pixel image from RGB to various alternative representations. The results indicate significant variation in computational cost among different color models. Table \ref{tab:processing_time} and Fig. \ref{fig:experiment1image} present the processing times for the conversion.

Among the tested color models, YIQ exhibits the fastest processing time (20.6 ms), followed by HSL (58.4 ms) and HSI (68.5 ms). In contrast, conversions to CIE Lab, CIE LUV, and CMYK demonstrate the highest computational cost, requiring 1.79 s, 1.15 s, and 922 ms, respectively. The CIE Lab transformation, known for its perceptual uniformity, incurs the highest processing time due to the complex nonlinear transformations involved. The relatively high computation time for CMYK can be attributed to its four-channel representation, which requires additional computations compared to three-channel models.
Overall, the results suggest that while perceptually uniform color spaces such as CIE Lab and CIE LUV provide advantages in color consistency, they impose substantial computational overhead. Conversely, models like YIQ and HSL offer significantly faster conversions, making them preferable for real-time applications where efficiency is critical.

These results offer insights into the computational efficiency of different color models, facilitating the selection of suitable models for various applications that require color transformations.

\begin{table*}
    \centering
    \begin{tabular}{|c|c|c|l|}
        \hline
        \textbf{Color Model} & \textbf{Processing Time (Mean ± Std. Dev.)} & \textbf{Relative Overhead (\%)} & \textbf{Speed Class} \\
        \hline
        CMY & 180.0 ms $\pm$ 28.3 ms & 10.06\% & Moderate\\
        CMYK & 922.0 ms $\pm$ 88.6 ms & 51.54\% & Very Slow\\
        CIE Lab & 1 790 ms $\pm$ 381.0 ms & 100\% & Very Slow\\
        CIE XYZ & 776.0 ms $\pm$ 31.0 ms & 43.35\% & Slow\\
        CIE LUV & 1150 ms $\pm$ 194.0 ms & 64.25\% & Very Slow\\
        HSL & 58.4 ms $\pm$ 1.16 ms & 3.26\% & Very Fast\\
        HSV & 127.0 ms $\pm$ 30.2 ms & 7.10\% & Moderate\\
        HSI & 68.5 ms $\pm$ 31.4 ms & 3.83\% & Very Fast\\
        YIQ & 20.6 ms $\pm$ 0.56 ms & 1.15\% & Very Fast\\
        YUV & 307.0 ms $\pm$ 44.1 ms & 17.15\% & Slow\\
        YCbCr & 113.0 ms $\pm$ 25.8 ms & 6.32\% & Fast\\
        \hline
    \end{tabular}
    \caption{Processing time for a 200×200 image (mean ± std. dev. of 7 runs, 10 loops each), normalized to the slowest model (CIE Lab = 100\%), and categorized into performance tiers from Very Fast to Very Slow. }
    \label{tab:processing_time}
\end{table*}

\subsection*{Experiment 2 - Intuitiveness}
\begin{figure}[t]
    \centering
    \includegraphics[width=\linewidth]{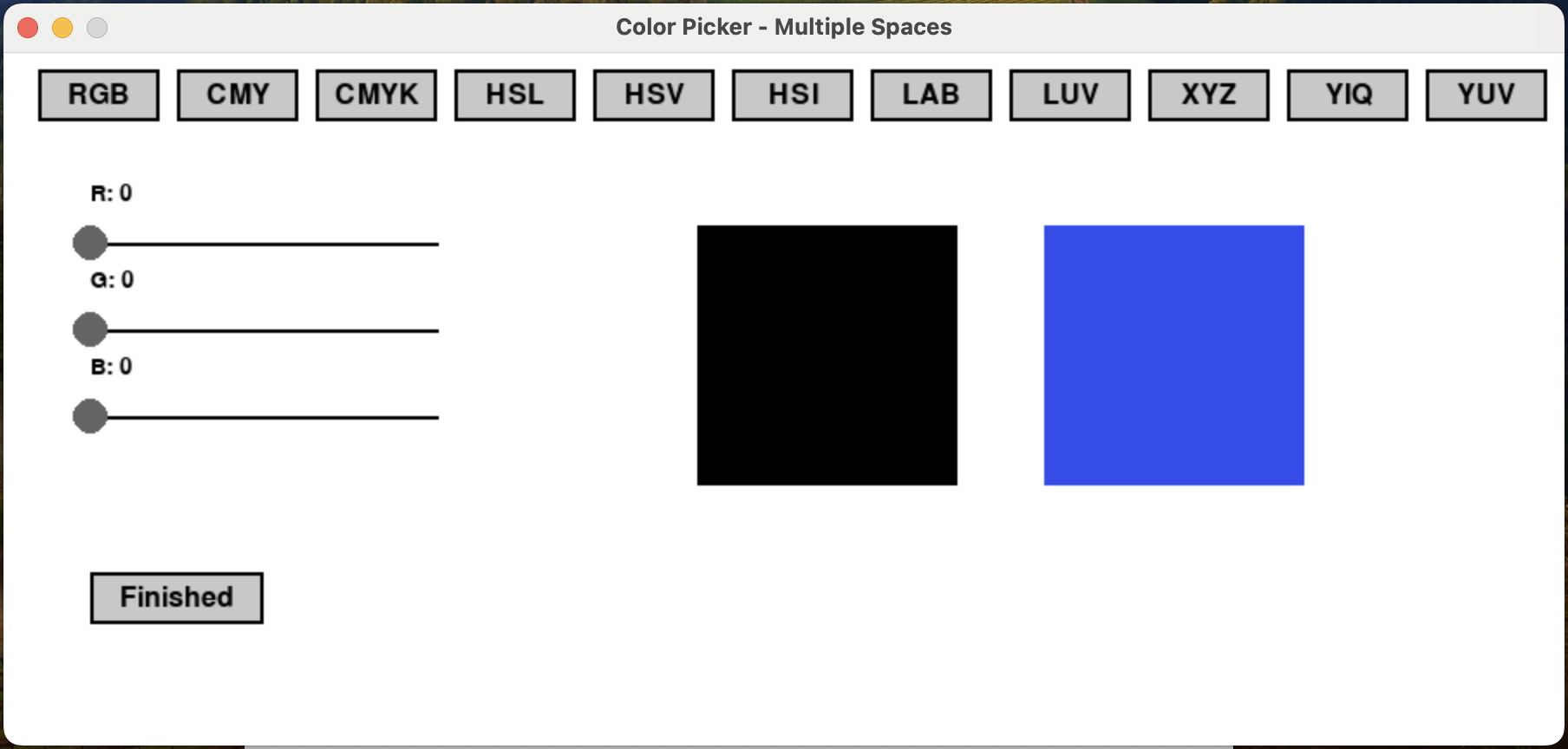}
    \caption{Experiment 2: Color Picker}
    \label{fig:picker}
\end{figure}

To evaluate the intuitive usability of different color models, we conducted an experiment involving seven participants. The Ethics Committee of Kazakh-British Technical University gave its approval to this research project, which included information gathering through human-centered experiment. All participants were domain experts with prior knowledge of color models and their underlying principles. Before the task, they were provided with a brief explanation of the component structure of each model used in the study. Furthermore, no personal identifying information was collected and each participant completed an informed consent form before starting the experiment.
For the experiment, we developed a custom software tool functioning as a color picker (Fig. \ref{fig:picker}). The interface consisted of slider-based controls, where each slider corresponded to one component of the selected color model. Ten color models were tested in total: RGB, CMY, CMYK, HSL, HSV, HSI, CIE LAB, CIE LUV, CIE XYZ, YIQ, YUV, and YCbCr.

In each trial, the program generated a target color, displayed as a filled square. Participants were asked to reproduce this target color by manipulating the sliders of the given color model. For example, in the RGB model, three sliders controlled the red, green, and blue components, and participants adjusted these until they believed the reproduced color matched the target. Once a participant confirmed their selection, the task proceeded to the next color model.

We measured the completion time for each trial, defined as the duration from the appearance of the target color to the participant’s confirmation of the match. This metric served as an indicator of the intuitive transparency of the respective color model, reflecting how easily participants could interpret and manipulate its components to achieve the desired result.

\begin{figure}[t]
    \centering
    \includegraphics[width=\linewidth]{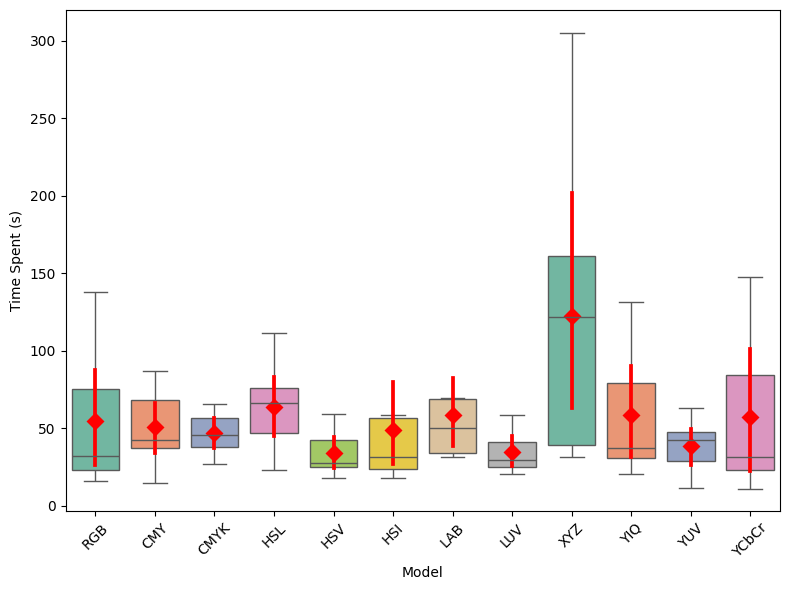}
    \caption{Experiment 2: Comparison of Time Spent}
    \label{fig:exp2_results}
\end{figure}

\begin{table}[ht!]
\centering
\caption{Categorization of color models by completion time. 
Lower average completion time indicates higher intuitiveness of the model. 
Models are grouped into three categories: high, medium, and low.}
\label{tab:intuitiveness}
\begin{tabular}{|l|c|c|c|}
\hline
\textbf{Mode} & \textbf{MeanTime (s)} & \textbf{Cluster} & \textbf{Intuitiveness} \\
\hline
CMY   & 50.58 & 1 & Medium \\
CMYK  & 46.69 & 1 & Medium \\
HSI   & 48.80 & 1 & Medium \\
HSL   & 63.88 & 1 & Medium \\
HSV   & 34.25 & 0 & High   \\
LAB   & 58.74 & 1 & Medium \\
LUV   & 34.37 & 0 & High   \\
RGB   & 54.69 & 1 & Medium \\
XYZ   & 122.71& 2 & Low   \\
YCbCr & 57.29 & 1 & Medium \\
YIQ   & 58.54 & 1 & Medium \\
YUV   & 38.63 & 0 & High   \\
\hline
\end{tabular}
\end{table}

The results demonstrated clear differences in intuitiveness across models. The HSI model yielded the shortest completion times, indicating that participants most rapidly understood the interplay of its components when matching colors. Conversely, the CIE XYZ model produced the longest times, suggesting it was the least intuitive among the tested models. A summary of the results is presented in Fig. \ref{fig:exp2_results}.

To identify natural groupings of color models in terms of completion time, k-means clustering (k=3) was applied to the mean values. This analysis yielded three categories of models' intuitiveness: high, medium, and low. As shown in Table~\ref{tab:intuitiveness}, models such as HSV, LUV, and YUV fall into the fast category, indicating higher intuitiveness for participants. In contrast, the XYZ model required substantially more time and was categorized as slow.

\subsection*{Comparison}

\begin{figure*}
    \centering
    \includegraphics[width=\linewidth]{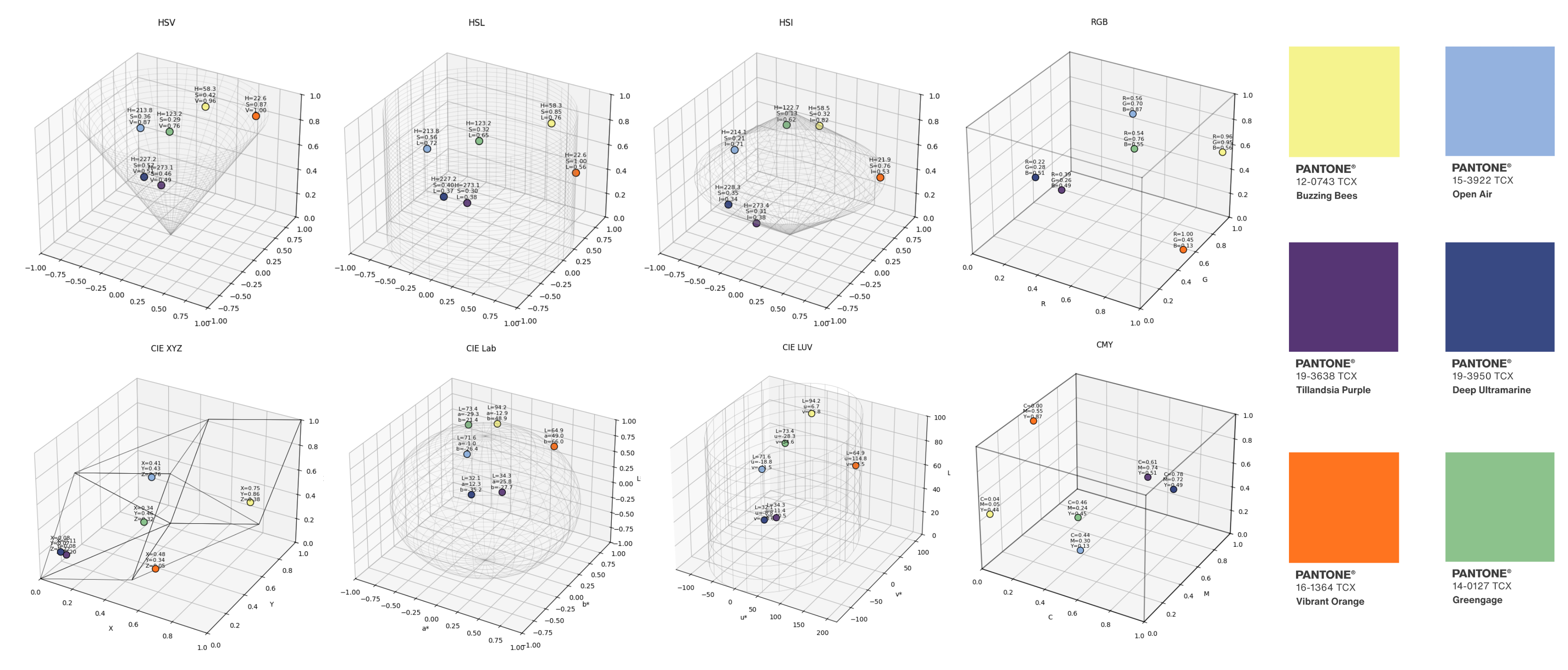}
    \caption{Representation of selected Pantone reference colors (“Buzzing Bees,” “Open Air,” “Tillandsia Purple,” “Deep Ultramarine,” “Vibrant Orange,” and “Greengage”) across different color models. The 3D plots illustrate the spatial distribution of the same colors in HSV, HSL, HSI, RGB, CMY, CIE XYZ, CIE Lab, and CIE LUV spaces, showing differences in geometry and perceptual uniformity among color models.}
    \label{fig:color_samples}
\end{figure*}

Fig. ~\ref{fig:color_samples} shows the distribution of Pantone reference colors across different color models.

The following aspects were chosen as critical when evaluating the effectiveness and suitability of color models for human perception-based applications:
\begin{itemize}
    \item \textbf{Attribute Dependency/Independency.} Considers whether attributes (e.g., hue, saturation, brightness) are independent or interdependent, important for flexibility in usage.
    \item \textbf{Device Dependency.} Addresses whether the color space is device-dependent or independent, this is important for preserving the consistency across devices \cite{Zhongjie_devicedepedency}. 
    \item \textbf{Chromatic-Achromatic Overlap.} Considers the separation of chromatic (color) and achromatic (brightness) components \cite{Egon_ChromaAchromaOverlap}. 
    \item \textbf{Preserving Chromatic Information.} Ensures the chromatic aspects of a color are preserved across operations or transformations.
    \item \textbf{Computation Speed. Integration with RGB.} Analyzes the computational efficiency, particularly for conversion to and from RGB. 
    \item \textbf{Application Area.} The range of fields and practical contexts where the color space can be effectively applied. 
    \item \textbf{User-Friendliness/Intuitiveness.} How easy it is for users to specify desired colors in the picker, for example, or interpret the color space.
    \item \textbf{Perceptual/Basic Uniformity.} Whether the color space provides perceptual uniformity (e.g., equal numeric changes correspond to equal perceptual differences) \cite{Ingmar_perceptualuniformity}. 
\end{itemize}

\begin{table*}[t]
    \centering
    \begin{tabularx}{\textwidth}{|X|X|X|X|X|}
        \hline
        \textbf{COLOR MODELS} & 
        \textbf{\centering Attribute \newline Dependency/ \newline Independency} & 
        \textbf{\centering Device \newline Dependency} & 
        \textbf{\centering Chromatic-\newline Achromatic Overlap} & 
        \textbf{\centering Preserving \newline Chromatic \newline Information} \\
        \hline
        Munsell & medium dependency & Independent & medium & n/a\\
        \hline
        RGB & high dependency & Dependent & high & medium (perceptual distortion) \\
        \hline
        CMY & high dependency & Dependent & high & low (printing distortion) \\
        \hline
        CMYK & high dependency & Dependent & high & low (printing distortion) \\
        \hline
        CIE Lab & Independent & Independent & low & high \\
        \hline
        CIE XYZ & high dependency & Independent & high & high \\
        \hline
        CIE LUV & Independent & Independent & low & high \\
        \hline
        HSL & medium dependency & Dependent & low & medium (hue distortion at particular saturation levels) \\
        \hline
        HSV & medium dependency & Dependent & low & medium (hue distortion at particular saturation levels) \\
        \hline
        HSI & medium dependency & Dependent & low & medium (hue distortion at particular saturation levels) \\
        \hline
        YIQ & medium dependency & Dependent \cite{Khalili2013} & low & low \\
        \hline
        YUV & medium dependency & Dependent & low & low \\
        \hline
        YCbCr & medium dependency & Dependent & low & low \\
        \hline
    \end{tabularx}
    \caption{Comparative overview of color models on different aspects (Part 1)}
    \label{tab:comparison_part_1}
\end{table*}
RGB was not included in the computation speed experiment because all images are natively stored in RGB format, eliminating the need for conversion. The Munsell system was also excluded since it cannot be directly converted to or from RGB, making it unsuitable for computational comparison in this context.

Table \ref{tab:comparison_part_1} summarizes structural aspects. Models with independent attributes, such as CIELAB and CIELUV, are more flexible for analysis, while device-dependent models like RGB and CMYK face cross-platform limitations. Clear chromatic–achromatic separation, seen in HSL or CIELUV, improves perceptual handling, and high stability in preserving chromatic information is characteristic of perceptually uniform models.

Table \ref{tab:comparison_part_2} presents practical aspects. HSV, HSL, and HSI show faster computation and greater intuitiveness but lower perceptual accuracy, whereas CIELAB and CIELUV achieve stronger uniformity at higher computational cost. Device-dependent models remain widely used in imaging and printing, while perceptually uniform spaces are better suited for visualization, color difference evaluation, and image processing tasks.

\begin{table*}
    \centering
    \begin{tabularx}{\textwidth}{|X|X|X|X|X|}
        \hline
        \textbf{COLOR MODELS} & 
        \textbf{\centering Computation Speed \ref{tab:processing_time}} & 
        \textbf{Application Area} &
        \textbf{\centering Intuitiveness} & 
        \textbf{\centering Perceptual Uniformity } \\
        \hline
        Munsell &  N/A & archaeology, geology, and anthropology for identifying colored surfaces and soils \cite{Stanco11} & N/A & N/A \\
        \hline
        RGB &  N/A & image processing systems, computer graphics, and multimedia applications & medium & low \cite{Shamoi2014Colorspace} \\
        \hline
        CMY & Moderate & printing & medium & low \\
        \hline
        CMYK & Very Slow & printing & medium & low \\
        \hline
        CIE Lab & Very Slow & color difference tasks, food coloring \cite{Jovanović2017Verification}, image processing for color enhancement \cite{Chiang2018Color}, surface grading \cite{López‐García2005Fast} & medium & high \cite{Paschos2001Perceptually} \\
        \hline
        CIE XYZ & Slow & image processing, computer vision, medical imaging, and color calibration, tasks requiring high color accuracy & low & low \cite{Paschos2001Perceptually} \\
        \hline
        CIE LUV & Very Slow & image compression, color segmentation, defect detection, remote sensing & high & high \\
        \hline
        HSL & Very Fast & image enhancement, segmentation tasks \cite{Garg2022, Kalist2015} & medium & - \\
        \hline
        HSV & Moderate & color-based segmentation, visibility enhancement, and traffic sign detection & high & - \\
        \hline
        HSI & Very Fast & image enhancement, video game design, emotional analysis \cite{Ma2018, Zhi2020} & medium & - \\
        \hline
        YIQ & Very Fast & \multirow{3}{3cm}{television broadcasting, digital watermarking} & medium & low \\ \cline{1-2} \cline{4-5}
        YUV & Slow &  & high & low \\ \cline{1-2} \cline{4-5}
        YCbCr & Fast &  & medium & low \\ \hline
    \end{tabularx}
    \caption{Comparative overview of color models on different aspects (Part 2)}
    \label{tab:comparison_part_2}
\end{table*}




\section*{Research Gaps and Future Work}

Despite the diversity and progress in color models and spaces, several open challenges remain, pointing to the need for further research in this area.

\begin{itemize}
    \item \textbf{Incomplete Perceptual Uniformity.} While models like CIELAB and CIEDE2000 aim to reflect human perceptual uniformity, they still exhibit inconsistencies in certain color regions. More research is needed to achieve perceptual linearity across the full visible spectrum.
  \item  \textbf{Underexplored Human-Centric Approaches.}
 Human color perception is inherently imprecise, context-dependent, and influenced by emotion, language, and culture. However, most conventional models treat color deterministically.  Few color models have been validated through user studies or cognitive experiments. Future work should focus on better reflecting subjective experience and including user-centered validation.

 \item \textbf{Task-Dependent Model Selection.}
There is no single color model that performs optimally across all applications. Practitioners must choose between models optimized for printing, display, compression, or perception. This calls for the development of hybrid or adaptive color models.

\item \textbf{Lack of Adaptivity to Lightning Conditions.}
Current models generally assume static lighting and viewing conditions. Incorporating illumination-aware or context-aware parameters into color models remains an open problem.
\end{itemize}

In summary, future research should aim to bridge the gap between computational efficiency, perceptual accuracy, and real-world applicability by developing adaptive color models that are validated through human-centered evaluations.
\section*{Conclusion}

This paper presents a comprehensive review of color models and color spaces, exploring their theoretical foundations, perceptual characteristics, computational properties, and practical applications in image processing.
From traditional models such as RGB and CMYK to perceptually uniform spaces like CIELAB and advanced fuzzy sets and logic-based approaches, we have highlighted the strengths, limitations, and use-case scenarios for each. This review involves user studies for experimental perceptual evaluation and human consistency, as well as experiments on computational efficiency. The experimental results demonstrate the trade-offs between simplicity and perceptual accuracy.
Traditional models, such as RGB and the HS family, are still widely used due to their speed, but modern applications require perceptually aligned models. Fuzzy color spaces, CIE LUV, and HSV are emerging as promising directions that bridge the gap between human perception and machine representation. This review serves as a reference for researchers and practitioners aiming to make informed decisions when choosing color models for specific tasks. The review has several limitations. It does not include a comparative evaluation of fuzzy versus classical color models in terms of perceptual accuracy. Additionally, it excludes performance assessment in key image processing tasks such as segmentation, enhancement, and retrieval.



\section*{Acknowledgment}
The authors would like to thank Yerdauit Torekhan for assisting in preparing the figures used in this paper.

\section*{Author contributions}
Muragul Muratbekova, Nuray Toganas, Ayan Igali, Maksat Shagyrov, Elnara Kadyrgali, Adilet Yerkin, and Pakizar Shamoi contributed equally to this work.

\section*{Funding}
This research has been funded by the Science Committee of the Ministry of Science and Higher Education of the Republic of Kazakhstan (Grant No. AP22786412).

\bibliography{export}

\end{document}